# Advancements in Point Cloud Data Augmentation for Deep Learning: A Survey

Qinfeng Zhu[a,b], Lei Fan[a,1], Ningxin Weng[a,c]

[a] Department of Civil Engineering, Xi'an Jiaotong-Liverpool University, Suzhou, 215123, China

[b] Department of Computer Science, University of Liverpool, Liverpool, L69 3BX, UK

[c] Department of Geography and Planning, University of Liverpool, Liverpool, L69 7ZT, UK

**Abstract**: Deep learning (DL) has become one of the mainstream and effective methods for point cloud analysis tasks such as detection, segmentation and classification. To reduce overfitting during training DL models and improve model performance especially when the amount and/or diversity of training data are limited, augmentation is often crucial. Although various point cloud data augmentation methods have been widely used in different point cloud processing tasks, there are currently no published systematic surveys or reviews of these methods. Therefore, this article surveys these methods, categorizing them into a taxonomy framework that comprises basic and specialized point cloud data augmentation methods. Through a comprehensive evaluation of these augmentation methods, this article identifies their potentials and limitations, serving as a useful reference for choosing appropriate augmentation methods. In addition, potential directions for future research are recommended. This survey contributes to providing a holistic overview of the current state of point cloud data augmentation, promoting its wider application and development.

## 1 Introduction

A point cloud comprises a collection of three-dimensional (3D) points in space. Point cloud data are typically acquired through sensors such as depth cameras, Light Detection and Ranging (LiDAR), and millimeter-wave radar. Point clouds have important and valuable applications in various application fields such as autonomous driving, 3D reconstruction, medical imaging, virtual reality, and augmented reality. In recent years, as a wide range of sensors has become more

[1] Corresponding author.

Email addresses: Qinfeng.Zhu21@student.xjtlu.edu.cn (Q. Zhu), Lei.Fan@xjtlu.edu.cn (L. Fan), Ningxin.Weng15@student.xjtlu.edu.cn (N. Weng)

accessible, researchers have shown an increasing interest in developing diverse techniques for the processing and analysis of point cloud data, particularly those grounded in deep learning (DL) methods [1].

In the field of DL, data augmentation is commonly used when available training datasets are limited. It involves performing a series of specific data operations to modify or expand the original data, thereby increasing the quantity and diversity of the dataset. Since well-augmented datasets contribute to improved robustness, enhanced generalization and reduced overfitting in trained networks [2], data augmentation is almost always deemed desirable when training DL networks. Comprehensive developments have been observed in both image data augmentation and text data augmentation, as reported in reviews or surveys [2-5].

Methods used for image augmentation provide useful insights into determining means of augmenting point cloud data [6, 7]. However, although certain image augmentation methods (e.g., translation, rotation, flipping and scaling) can be readily applied to augment point cloud data, their performance may vary from two-dimensional (2D) images to 3D point clouds. The distinctive characteristics of point cloud data often require augmentation methods specifically designed or tailored for this type of data. In addition, unlike image-based DL training where various image benchmarks have extensive training samples, there are only a limited number of public benchmark point cloud datasets, typically characterized by limited class labels and data diversity. This highlights the greater significance of data augmentation for point cloud data compared to image data. Consequently, extensive research on point cloud data augmentation has been carried out in recent years.

In numerous recently published research papers on point cloud processing tasks, researchers have explored various methods of augmenting point cloud data. This wide range of methods poses a challenge for researchers in selecting suitable methods. Therefore, there is significant value in systematically surveying these methods and categorizing them into different groups. In their survey of label-efficient learning of point clouds, Xiao et al. [8] broadly categorized conventional point cloud augmentation into intra-domain and inter-domain augmentation, according to whether additional data modalities are used. While this classification is simple, it does not represent the diversity of point cloud augmentation methods.

In this article, we present a comprehensive survey of augmentation methods for point cloud data. Based on our survey, we propose a taxonomy of these augmentation methods, illustrated in Figure 1. Augmentation methods are broadly divided into two main categories: basic and specialized point cloud augmentation, which are similar to typical categorization of image augmentation (e.g., [2, 4, 5]). Basic point cloud augmentation refers to approaches that are not only simple in concept but also versatile in diverse tasks and application contexts, evidenced by their widespread utilization in combination with one another in the surveyed literature. Specialized point cloud augmentation refers to methods that are typically developed to tackle specific challenges or address particular application contexts. In most cases, specialized augmentation tends to be more computationally complex than basic augmentation, depending on the implementation details of augmentation methods. The subcategories in our proposed taxonomy represent an inclusive summary of diverse methods that have been used for point cloud data augmentation in the literature or have potential of being employed for augmentation. These subcategories are elaborated in Sections 2 and 3.

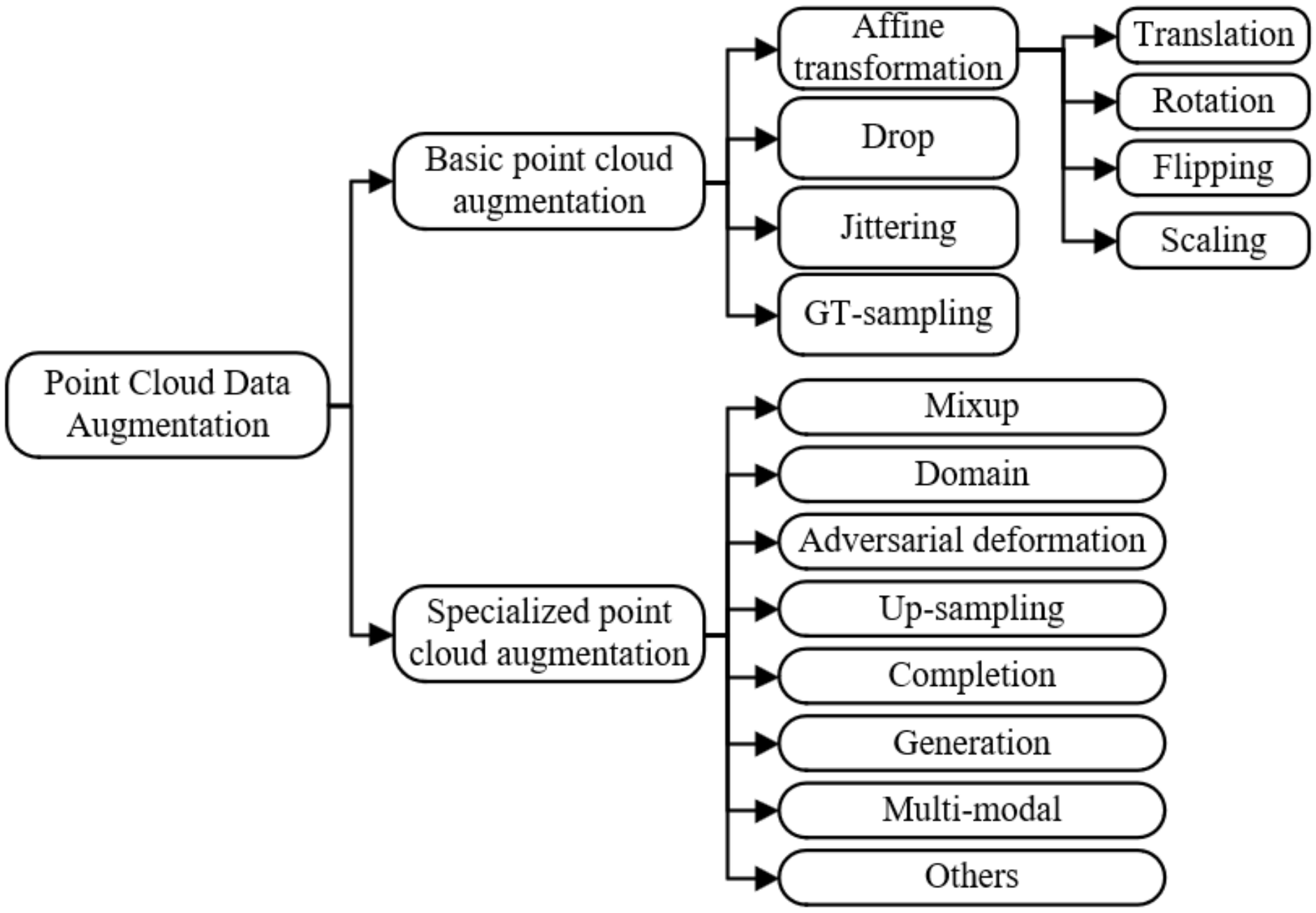

Figure 1: Taxonomy of point cloud data augmentation methods.

The main contributions of our article are as follows:

- To the best of our knowledge, this is the *first* comprehensive survey of methods for augmenting point cloud data augmentation, which covers recent advancements in point cloud data augmentation. According to the complexity and nature of augmentation operations, we propose a taxonomy of point cloud data augmentation methods.

- This study summarizes various point cloud data augmentation methods, discusses their applications in typical point cloud processing tasks such as detection, segmentation and classification, and provides suggestions for potential future research.

The rest of this article is structured as follows. Section 2 introduces basic point cloud augmentation methods. Section 3 demonstrates specialized point cloud augmentation methods. Section 4 discusses briefly the application scenarios of basic and specialized augmentations, addresses existing limitations associated with current point cloud data augmentation, and proposes potential research directions. Finally, we conclude our survey in Section 5.

# 2 Basic point cloud augmentation

## 2.1 Typical basic operations

### 2.1.1 Affine transformation

Affine transformation involves the transformation of an affine space, which preserves collinearity and distance ratios. In the context of image data augmentation, commonly used affine transformation methods include scaling, translation, rotation, reflection and shear. Similarly, affine transformation can also be applied to point cloud data augmentation. Typical methods include translation, rotation, flipping and scaling, and have been widely used to generate additional new training data. These operations can be applied to the entire set of point cloud data, to selected instances (an instance refers to a semantic object such as the vehicle shown in Figure 2(a)) in the point cloud data using specific strategies, or to specific part(s) of selected instances. Data augmented by affine transformations may face the problem of information loss or semantic unreasonableness, which are explained in the following specific operations.

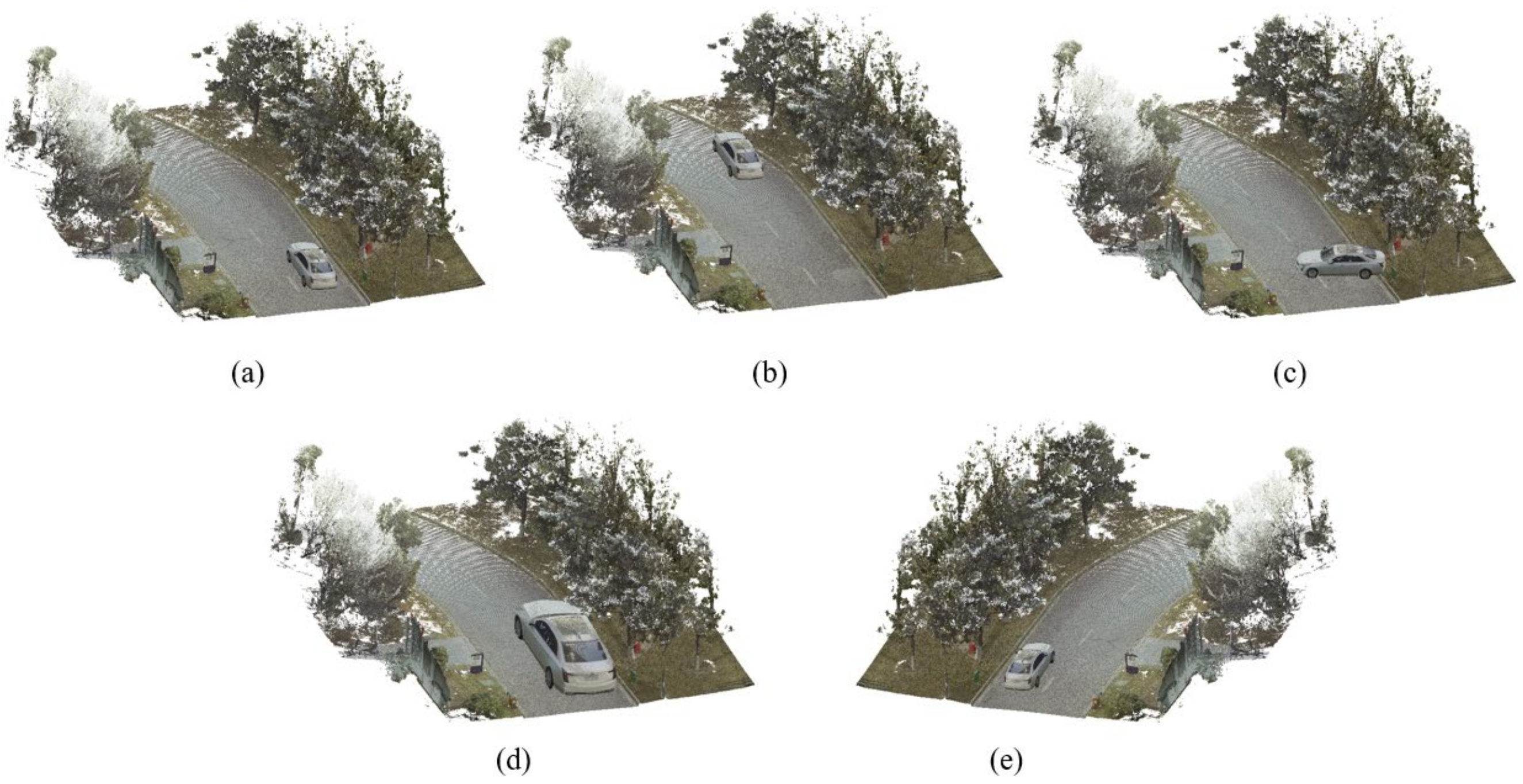

Figure 2: Examples of point cloud data augmentation through affine transformation: (a) Original point cloud data, (b) Translated vehicle, (c) Rotated vehicle, (d) Scaled vehicle, (e) Flipped scene.

*Translation* represents the operation of moving selected point cloud data by a specific distance and direction, as shown in Figure 2(b). This augmentation helps DL models to better learn instances at various locations, reducing models' sensitivity to spatial variability of instances within the scene. However, careful selection of translation ranges and directions is important. For example, translation may lead to occlusions or result in instances appearing in semantically inappropriate locations, such as when cars overlap with buildings, which deviates from the reasonable scene. Additionally, extensive translations can shift instances outside the targeted scene, causing a loss of information.

*Rotation* represents the operation of rotating selected point cloud data according to a specified direction and angle, as shown in Figure 2(c). This augmentation can be used to simulate different object orientations or sensor viewpoints, enhancing DL models to handle instances' pose variations. In datasets where instances' posture is predominantly upright [9], rotation augmentation should be considered to account for real-world scenarios involving sensor tilt and instance rotation. In rotation augmentation, a diverse set of rotations may be required to adequately represent different instance orientations, especially uncommon ones. However, this augmentation may not be suitable for cases where the absolute object orientation is critical. Additionally, large rotations may distort instances, potentially affecting models' ability to learn meaningful features.

*Scaling* involves performing a scale transformation on selected point cloud data, as shown in Figure 2(d). This augmentation can simulate various object sizes, enhancing DL models' adaptability to scale variations. Due to variations in networks' receptive fields, the original input data may not be at the optimal scale for a network. By employing scaling operations, the network can process point cloud data of varying input sizes. Numerous studies have highlighted the effectiveness of multi-scale training [10], especially for small target perception [11]. However, attention should be paid to choosing appropriate scales to avoid introducing unrealistic geometry that may affect the spatial relationships between data points. In addition, selection of scaling ranges should avoid oversampling or undersampling instances or regions in the point cloud data since scaling will increase or decrease point data densities.

*Flipping* represents the operation of flipping selected point cloud data along a specified axis, as shown in Figure 2(e). This method enhances the generalization ability of DL models to instance orientations and symmetry features. In instance-level datasets, horizontal flipping and vertical flipping are both commonly used. However, in scene-level datasets, vertical flipping does not contribute to improved model discriminability [12]. For example, in the autonomous driving dataset nuScenes [13], only horizontal flipping was used to prevent semantic ambiguity. Vertically flipped instances of people and cars in the scene can be semantically unrealistic, potentially affecting model performance.

### 2.1.2 Drop

Drop refers to discarding some data points in point cloud data, as shown in Figure 3. The selection of points to remove is determined by specific strategies formulated by researchers. The discarded points can be part(s) of the whole point cloud data or randomly selected points in a scene. Drop augmentation helps DL models become more robust to missing or incomplete data representing occluded or partially visible scenes. It may also prevent DL models from relying too heavily on specific data points in the training dataset. However, losing excessive or key point cloud information may lead to unrealistic representation of real-world objects in the training data, and affect the training of DL models, particularly when large data densities are critical or objects are small.

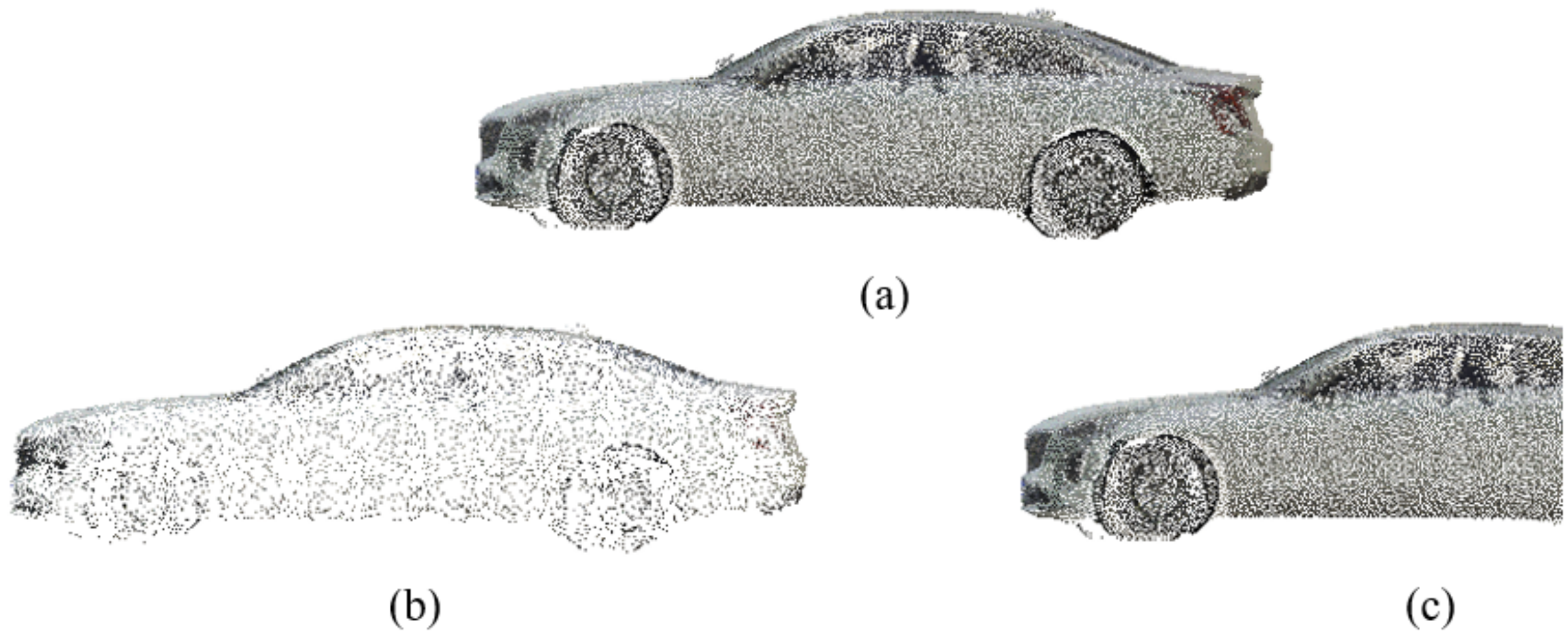

Figure 3: Examples of point augmentation through drop augmentation: (a) Original point cloud data, (b) Augmented point cloud with randomly dropped points, (c) Augmented point cloud with a dropped part.

A simple strategy for dropping points in an object or scene involves the random removal of points. While this strategy is straightforward, it may cause the loss of key points for subsequent point cloud processing tasks. Two example methods of this nature are exemplified in the following. In one method, Zhang et al. [14] generated a specific number and size of cuboids in the point cloud data, and randomly discarded points within a specific proportion of cuboids. This method forces the model to predict the discarded information through other available geometric features. Similarly, IPC-Net [15] uses the idea of random erase to simulate data loss. This method randomly selects a location in the 2D view of a point cloud instance as the centroid, and then discards all points inside a circle centered at the centroid with a random radius.

To honor key points or semantic information in point cloud data, more sophisticated drop algorithms have also been developed. PointDrop [16] uses an adversarial learning method to identify key points in the point cloud data and drop them to obtain more challenging sparse samples for training. In 3DEPS [17], the point cloud data of plants is differentiated into edge and non-edge regions. To augment data while preserving plant structure, a selective drop strategy is applied: edges undergo minimal point drop to maintain structural integrity, while non-edge areas, such as interiors between leaves and branches, experience a more substantial drop. This approach aims to safeguard critical edge information vital for recognizing plant morphology, ensuring that augmented data retains essential structural details. This type of approaches is particularly beneficial in applications where complex edge structures are essential for subsequent point cloud processing tasks.

### 2.1.3 Jittering

Jittering refers to applying small perturbations or noise to the positions of individual points in a point cloud, as shown in Figure 4. First, the magnitude of perturbation is determined by users. Second, a perturbation vector is generated for each point according to the perturbation magnitude determined. This perturbation vector is usually generated using random sampling or random number methods [18, 19]. The last step is to perturb the points. In this step, each point in the point cloud is perturbed by its corresponding perturbation vector and transformed into a perturbed point cloud.

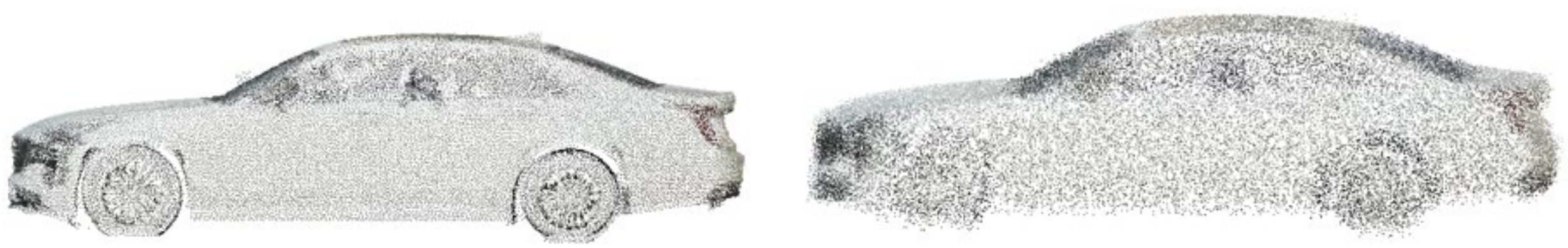

(a) (b)

Figure 4: Example of jittering augmentation: (a) Original point cloud data, (b) Jittered point cloud data.

Jittering can simulate small spatial variations induced by measurement errors and enhance a DL model's resistance to noise, especially at object edges. However, excessive perturbation may also destroy the local structural features of the point cloud, thereby affecting the performance of the model during training. The effectiveness of jittering is likely to vary across different tasks, task-specific experimentation is necessary to determine its impact on model performance. According to Table 1, jittering is frequently utilized in tasks such as segmentation and classification, where the introduced variability can enhance model robustness. However, its application in detection tasks is rare, where preserving the precise spatial relationships between points is often deemed more critical.

### 2.1.4 GT-sampling

In point cloud datasets at the scene level, such as those for outdoor autonomous driving scenarios, the availability of labeled instances is usually limited. In such cases, GT-sampling emerges as a simple and efficient data augmentation method. GT-sampling refers to the operation of adding labeled instances to a training dataset, with labeled ground-truth instances being sourced from the same training dataset or other datasets. GT-sampling is often applicable to scene-level point cloud datasets, and is typically not considered for instance-level point cloud datasets such as ShapeNet [20].

GT-sampling is one of the most commonly used methods for point cloud data augmentation due to its simplicity and effectiveness. It is often used in combination with other affine transformation operations for each added instance. The concept was first proposed in SECOND [21], where labeled instances were randomly selected and added to the training set, followed by the removal of instances that collided with existing objects. Subsequent GT-sampling methods [22-28] used the same or similar strategies to build on or refine the approach introduced by SECOND.

Instances added through GT-sampling may encounter semantic issues, leading to inappropriate instances being placed in inappropriate locations within the targeted scene environment. Examples of such issues include scenarios where an added car is placed amidst trees or collides with a building, as illustrated in Figure 5. These issues are common in various studies [29]. In response, researchers have improved GT-sampling for more realistic semantic positioning. For instance, Fast point R-CNN [30] involves copying the point cloud data from a small area surrounding the instance to be replicated, preserving the contextual information for the added instance. However, the added instances may still not align with the targeted scene environment, especially when scenes are complex. To improve the semantic plausibility, Hu et al. proposed CA-AUG [31], which identifies semantically reasonable locations within the point cloud data for placing instances and adds labeled instances in these reasonable locations. Similarly, 3D-VDNet [32] searches reasonable locations to avoid data collision and places added instances in these identified places.

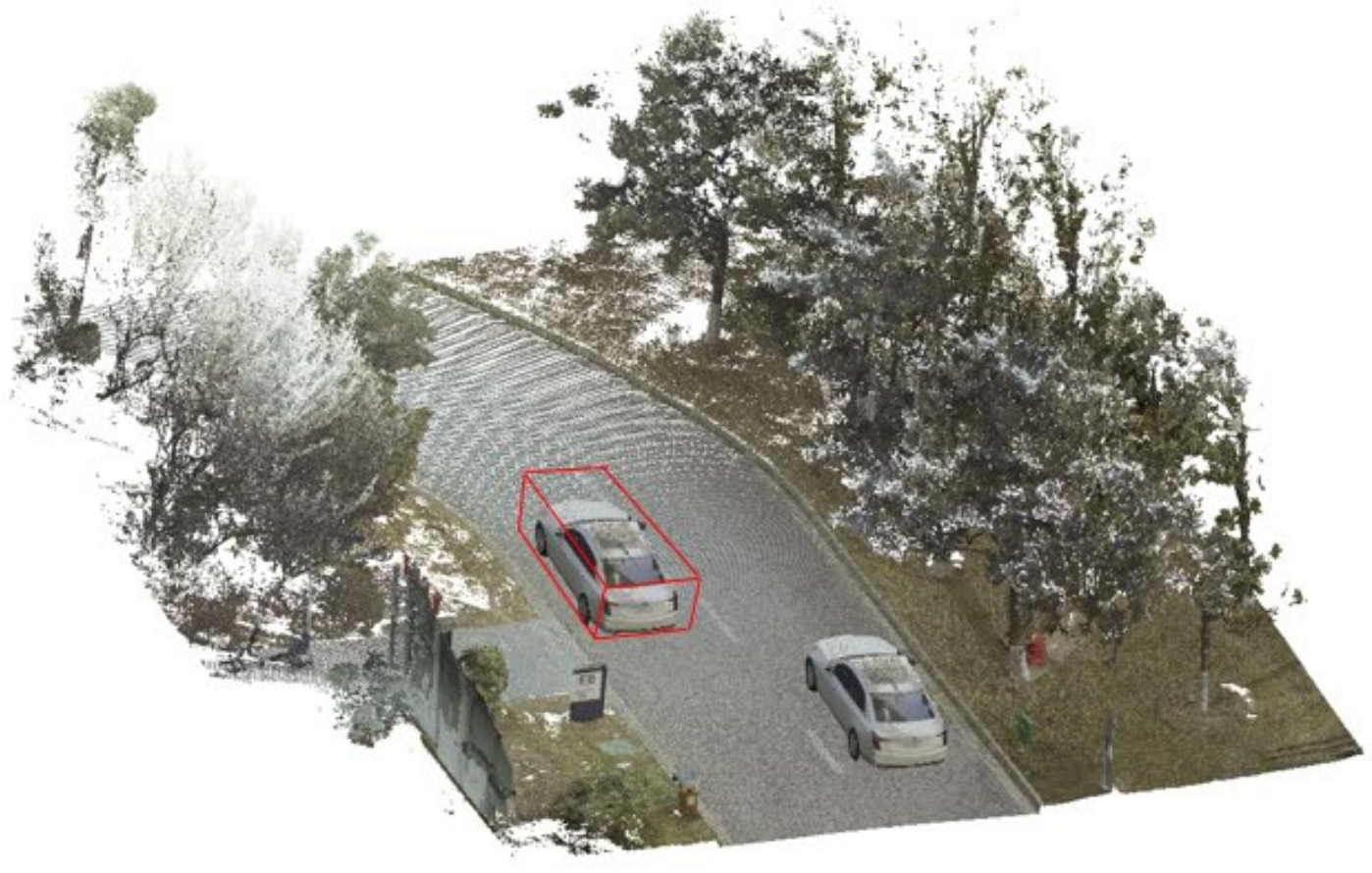

(a)

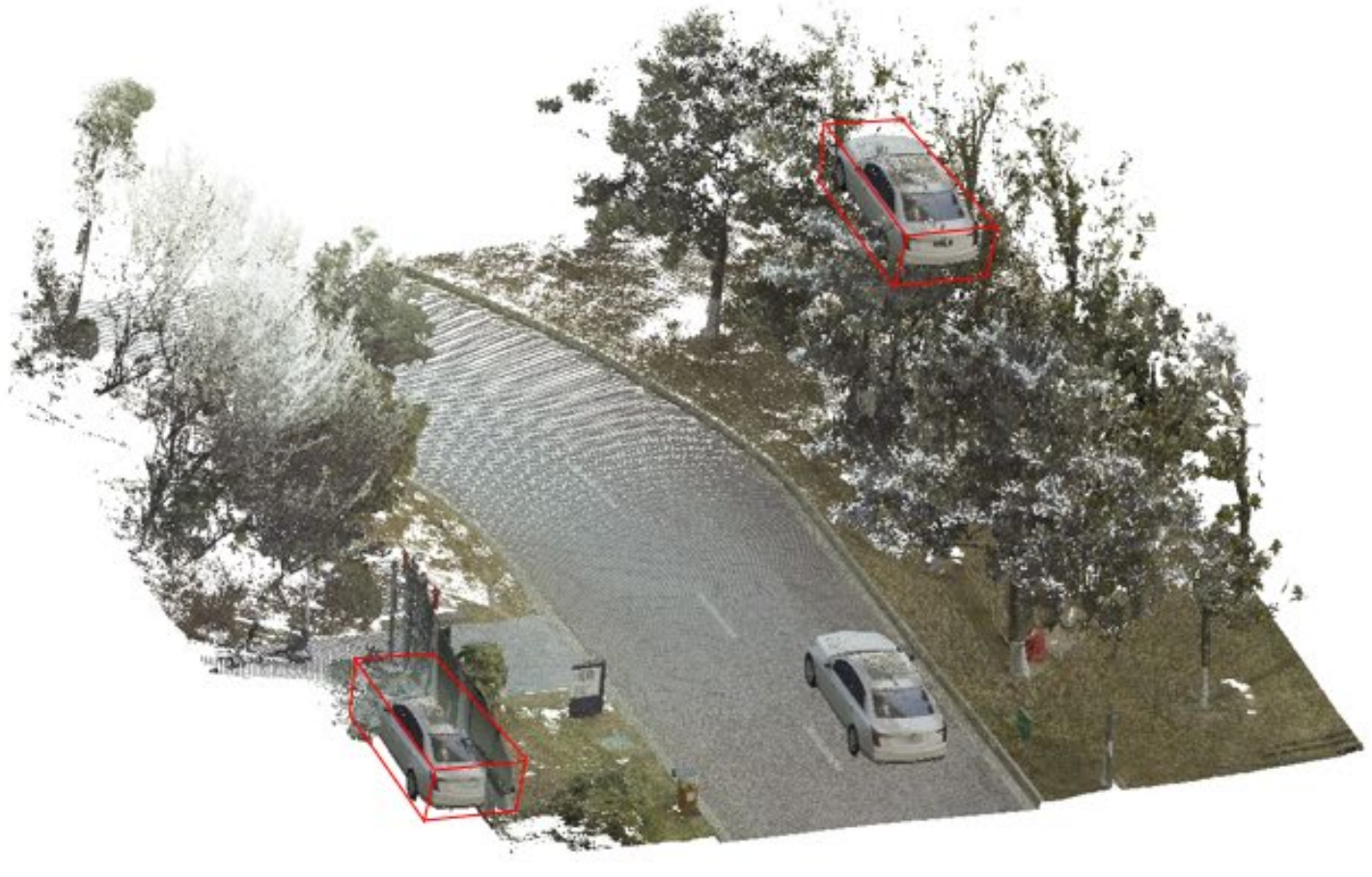

(b)

Figure 5: (a) Semantically reasonable GT-sampling, with the added vehicle in the red voxel. (b) Semantically unreasonable GT-sampling, with one added vehicle inside the building wall and the other into trees.

Some researchers have explored means of enhancing the GT-sampling strategy to address the issue of class imbalance. For example, RangeFormer [18] duplicates ground truth objects from the classes with fewer instances to targeted training scenes, and performs translation to change their positions. Zhu et al. [33] proposed dataset sampling (DS) to tackle class imbalance in autonomous driving scenes. DS selects instances in the minority classes and adds them to the training set to create a more balanced instance distribution. However, neither of these approaches considers the semantic fitness of replicated instances to targeted scenes. Therefore, it is beneficial to consider combining augmentation strategies developed for semantic plausibility and class imbalance during augmentation.

Another typical challenge associated with numerous GT-sampling methods is their inability to replicate data occlusion occurrences resulting from added instances, commonly encountered during data acquisition by LiDAR sensors. Addressing this issue, PointAugmenting [34] not only avoids collisions between added instances and the targeted scene, but also filters out points under occlusion by retaining those closer to the sensor. Similarly, Hasecke et al. [35] designed a competition mechanism for point injection, keeping points closer to the sensor and deleting points farther away in cases of occlusion. These methods offer the advantage of inserting labeled instances with realistic occlusion into the augmented dataset.

## 2.2 Global and local point cloud augmentation

Point cloud data augmentation can be applied globally or locally. Global augmentation augments point cloud data in an entire scene. In both indoor and outdoor point cloud scenes, affine transformations, drop, and jittering has been widely adopted by researchers to globally augment an entire scene [36]. For instance, in the context of autonomous driving, Kong et al. [18] employed scaling, rotation, flipping, drop, and jittering to globally augment the whole scene. Lai et al. [37] used scaling, rotation, drop, and jittering to globally augment an entire indoor scene. In contrast, local augmentation targets specific part(s) or instance(s) within the whole point cloud, keeping the overall scene unchanged. *Patch-based augmentation* is a special case of local augmentation, which augments specific part(s) of an instance through various transformation operations. This aids DL models in better learning local features of instances for improved model performance in point cloud processing and analysis. In addition to typical basic operations, swap operations [38, 39] are sometimes used in patch-based augmentation, in which all points of two augmented instance parts are swapped.

There are various strategies for selecting instance patches to be augmented. For instance, SE-SSD [39] partitions a car instance into six pyramids and individually augments the point cloud data in each pyramid using operations like random drop and swap. While this method can strengthen the model's perception of the car’s shape, it may not be suitable for other types of objects. Likewise, PA-AUG [38] divides a given instance into 4 or 8 cuboids and applies several augmentation operations to point cloud data in each cuboid, which is applicable to the perception of various object classes. PointWOLF [40] performs smooth non-rigid deformations on parts of an instance, by selecting multiple anchor points through farthest point sampling and then deforming local areas within the range of the anchor points. This approach generates visually realistic and diverse new training data, but may not be suitable for certain classes of regularly shaped objects, such as cars. PatchAugment [41] uses k nearest neighbor or sphere querying [42] to divide an instance into numerous grouped points, and then performs a combination of basic operations on these groups together.

## 2.3 Application of basic operations

Basic point cloud augmentation has witnessed significant development in recent years, and some representative methods in this category are shown in Table 1. This table provides diverse information about different methods, including publication details, authors' designation for their augmentation method, augmentation operations employed, an indication of whether auto-optimization or patch-based augmentation is used, typical point cloud data processing tasks, relevant benchmark datasets, and application scenarios.

In certain point cloud processing tasks, rotation augmentation [43-48] and drop augmentation [14-17, 49] are sometimes used independently, but more commonly in combination with other basic operations, as shown in Table 1. GT-sampling, in conjunction with other affine transformation operations, is frequently used for the detection task, occasionally for segmentation, but rarely for classification. For classification, commonly combined basic operations include translation, rotation, scaling, jittering and drop. Combinations of basic operations are observed in many other studies (e.g., [50-58]). The selection of individual basic operations to combine is often based on extensive trial experiments or practical experience. In some studies [36, 59], the impacts of different combinations of basic operations on the performance of point cloud processing task(s) were tested and found to be ambiguous. Consequently, researchers often need to conduct numerous tests in a wide search space to identify the optimal combination for a specific application.

Table 1: Representative basic point cloud augmentation methods. 'GT', 'T', 'R', 'S', 'F', 'D', and 'J' represent GT-sampling, Translation, Rotation, Scaling, Flipping, Drop, Jittering, respectively.

| Year | Month | Authors | Name | Operations | Strategy | Task | Benchmark Dataset | Application |
|---|---|---|---|---|---|---|---|---|
| 2018 | Oct | Yan et al. [21] | - | GT+T+R+S | - | Detection | KITTI | Outdoor Autonomous Driving |
| 2019 | Apr | Griffiths et al. [48] | - | R | - | Segmentation | ScanNet, Semantic3D | Indoor and Outdoor Scene Understanding |
| 2019 | Jun | Choy et al. [60] | - | T+R+S+D+J | - | Segmentation | ScanNet, S3DIS, RueMonge 2014, Synthia 4D | Indoor and Outdoor Scene Understanding |
| 2019 | Aug | Chen et al. [30] | - | GT+T+R+S+F | - | Detection | KITTI | Outdoor Autonomous Driving |
| 2019 | Aug | Zhu et al. [33] | GT-AUG | GT+T+R+S+F | - | Detection | nuScenes | Outdoor Autonomous Driving |
| 2019 | Aug | Liu et al. [45] | - | R | - | Classification, Segmentation, Scene Flow Estimation | Synthia, KITTI scene flow, MSRAction3D, FlyingThings3D | Outdoor Dynamic Scene Understanding |
| 2020 | Feb | Li et al. [61] | PointAugment | T+R+S+J | Auto | Classification | ModelNet40, SHREC16 | 3D Object Understanding |
| 2020 | Aug | Cheng et al. [62] | PPBA | T+R+S+F+D | Auto | Detection | KITTI, Waymo | Outdoor Autonomous Driving |
| 2021 | Jan | Ma et al. [16] | PointDrop | D | - | Detection | KITTI | Outdoor Autonomous Driving |
| 2021 | Feb | Wang et al. [34] | PointAugmenting | GT+T+R+S+F | - | Detection | nuScenes, Waymo | Outdoor Autonomous Driving |
| 2021 | Feb | Zheng et al. [39] | - | T+R+S+F+D | Patch | Detection | KITTI | Outdoor Autonomous Driving |
| 2021 | Jun | Wang et al. [63] | - | GT+R+S+F | - | Detection | ScanNet, SUN-RGBD, KITTI | Indoor and Outdoor Scene Understanding |
| 2021 | Jun | Zhao et al. [52] | - | R+J | - | Segmentation | ScanNet, S3DIS | Indoor Scene Understanding |
| 2021 | Aug | Kim et al. [40] | PointWOLF | T+R+S+D+J | Auto+ Patch | Classification, Segmentation | ModelNet40, ScanObjectNN, ShapeNetPart | 3D Object Understanding |
| 2021 | Aug | Sheshappanavar et al. [41] | PatchAugment | T+R+S+D+J | Patch | Classification | ModelNet40, ModelNet10, SHREC16, ScanObjectNN | 3D Object Understanding |
| 2021 | Aug | Zhang et al. [14] | - | D | - | Classification, Segmentation, Detection | ScanNet, SUNGRBD, Matterport3D, S3DIS | Any |
| 2021 | Oct | Choi et al. [38] | - | T+R+S+F+D | Patch | Detection | KITTI | Outdoor Autonomous Driving |
| 2021 | Dec | Zhang et al. [19] | AdaPC | T+R+S+J | Auto | Classification | ModelNet40 | 3D Object Understanding |
| 2022 | Feb | Li et al. [17] | 3DEPS | D | - | Segmentation | 3D Plant Dataset | 3D Plant |
| 2022 | Jun | Hasecke et al. [35] | - | T+S+F+D | - | Segmentation | SemanticKITTI | Outdoor Autonomous Driving |
| 2022 | Jun | Lai et al. [37] | - | R+S+J | - | Segmentation | ScanNet, S3DIS, ShapeNetPart | Indoor Scene Understanding |
| 2022 | Nov | Xiao et al. [32] | - | GT+R+S+F | - | Detection | KITTI | Outdoor Autonomous Driving |
| 2023 | Mar | He et al. [15] | - | D | - | Classification | ModelNet40, ScanObjectNN | 3D Object Understanding |
| 2023 | May | Leng et al. [64] | LidarAugment | T+R+S+F+D | Auto | Detection | Waymo, nuScenes | Outdoor Autonomous Driving |
| 2023 | Oct | Hu et al. [31] | CA-AUG | GT+T+R+S+F | - | Detection | KITTI | Outdoor Autonomous Driving |
| 2023 | Oct | Kong et al. [18] | - | GT+T+R+S+F+D+J | - | Segmentation | SemanticKITTI, nuScenes, ScribbleKITTI | Outdoor Autonomous Driving |

However, researchers have also developed specific methods to automatically obtain the optimal combination of basic operations and their associated parameters through the training of neural networks. These methods are referred to as *Auto optimization*, as illustrated in the flow chart in Figure 6. Several representative auto optimization methods are elaborated in the following. PointAugment [61] is an end-to-end network capable of automatically selecting the best combination of augmentation operations from rotation, scaling, translation and jittering. It uses an augmenter to augment the input samples to obtain augmented samples, and then trains both the input samples and the augmented samples in the classifier, and feedbacks the result to the augmenter. Compared to PointAugment, Zhang et al. [19] proposed an automatic augmentation method named AdaPC, which is based on a two-layer optimization framework. AdaPC employs basic operations (i.e., scaling, rotation, and translation) in the augmenter and augments point cloud data with the minimized validation loss. This method was reported to exhibit generalization to more challenging scenes. While PointAugment and AdaPC were designed to automatically select the optimal combination of basic operations for point cloud classification, their applications to other tasks need further validation. Progressive Population Based Augmentation (PPBA) [62] uses four basic operations in training. It uses two basic operations for data augmentation in each iteration, narrows the search space through continuous iterations, and finds the best parameters in iterations. This method has shown effectiveness in improving model performance in detection and is deemed to have potential for extension to other point cloud processing tasks. Augmentation Tuning (AugTune) [40], from PointWOLF introduced in Section 2.2, is also an effective auto optimization method and can adaptively control the parameters of point cloud augmentation during training to generate targeted augmented data.

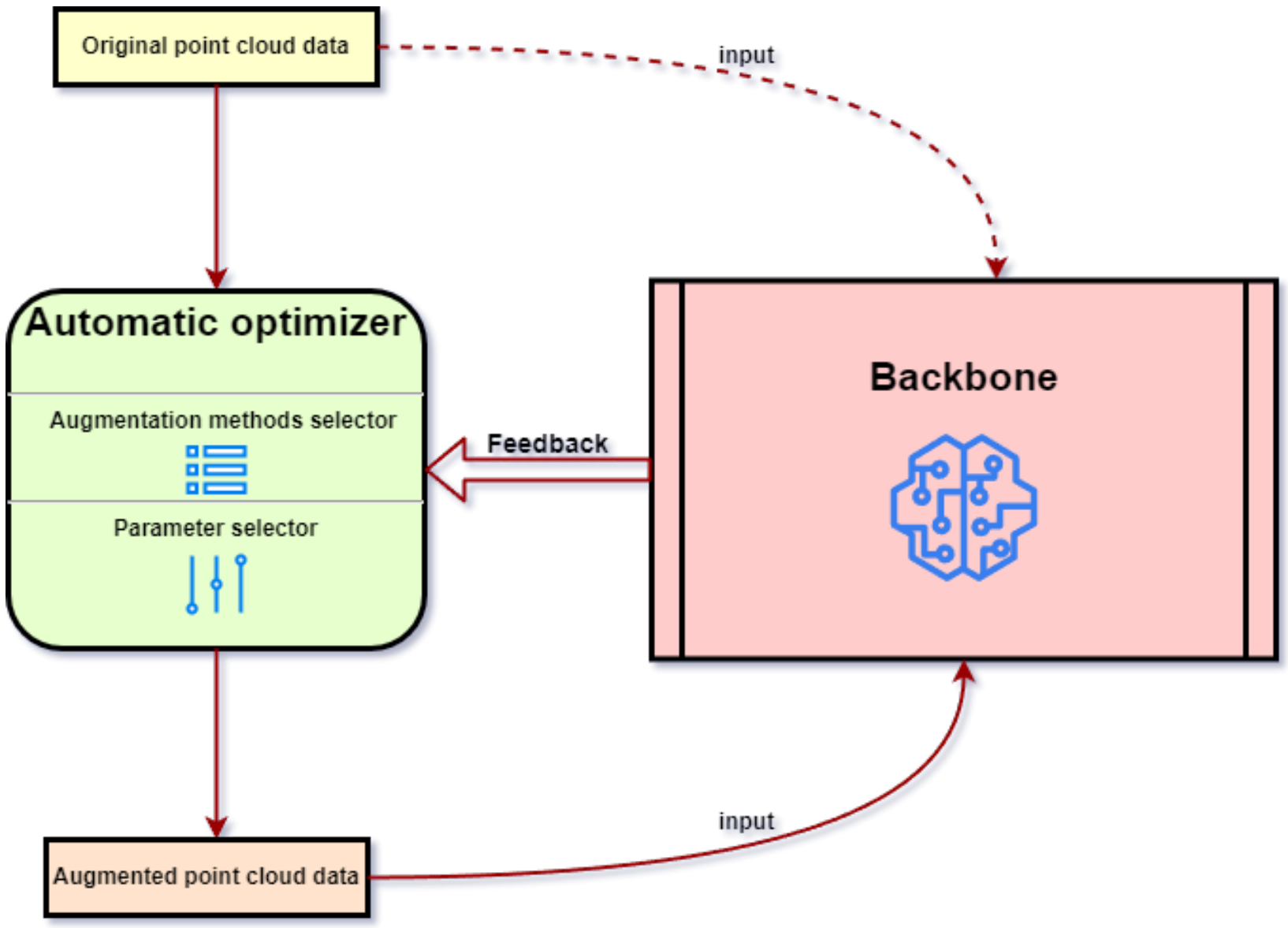

Figure 6: Typical process of auto optimization. The original point cloud data are augmented using an automatic optimizer. The resulting augmented point cloud data (and optionally the original point cloud data) are fed into a backbone network for training. The backbone network will convey the training results to the automatic optimizer, facilitating automatic optimizer to adjust the strategy and to determine the optimal augmentation and the parameter selection.

The aforementioned auto optimization methods consider a limited set of basic operations. This limitation arises from the challenge of handling a large search space when finding optimal parameters for many basic operations in auto optimization. LidarAugment [64] addresses this issue by using factorization and alignment search space, significantly reducing hyperparameters and computational complexity. It automatically finds the optimal augmentation scheme among ten augmentation operations.

Although auto optimization can adaptively select augmentation methods and parameters, it may produce unintended side effects that contradict the overall goal of data augmentation, affecting the model's ability to generalize to unseen data [65, 66]. During the iterative process of auto optimization, to fill the latent feature distribution, auto optimization strategies may select data augmentation patterns that are highly tailored to the training data, leading to the generation of augmented samples significantly different from those not included in the training data. Consequently, the model may overlook more generalized patterns in the original data, which detrimentally affects its generalization ability. Future research can focus on ensuring auto optimization avoids augmentations that deviate significantly from real-world scenarios, to reduce the risk of overfitting. Potential measures include assessing the generalizability of augmented data or restricting the types of augmentation operations utilized.

## 3 Specialized point cloud augmentation

Specialized methods are typically aimed at addressing specific challenges or application contexts. Table 2 outlines the development of representative specialized augmentation methods, providing diverse information, including publication details, authors' designation for their augmentation method, augmentation operations employed, typical point cloud data processing tasks, relevant benchmark datasets, and application scenarios.

Table 2: Representative specialized point cloud augmentation methods.

| Year | Month | Authors | Name | Augmentation | Task | Benchmark Dataset | Application |
|---|---|---|---|---|---|---|---|
| 2019 | Jan | Goodin et al. [67] | - | Domain | - | - | Outdoor Autonomous Driving |
| 2019 | Jun | Sallab et al. [68] | - | Domain | - | - | Outdoor Autonomous Driving |
| 2020 | Dec | Chen et al. [69] | PointMixup | Mixup | Classification | ModelNet40, ScanObjectNN | 3D Object Understanding |
| 2021 | Feb | Harris et al. [70] | Fmix | Mixup | Classification | ModelNet10 | 3D Object Understanding |
| 2021 | Jun | Lee et al. [71] | RSMix | Mixup | Classification | ModelNet40, ModelNet10 | 3D Object Understanding |
| 2021 | Jun | Gong et al. [72] | Maxup | Adversarial Deformation | Classification | ModelNet40 | 3D Object Understanding |
| 2021 | Jun | Fang et al. [73] | LiDAR-Aug | Multi-modal | Detection | KITTI | Outdoor Autonomous Driving |
| 2021 | Jul | Kilic et al. [74] | LISA | Domain | Detection | KITTI, Waymo | Outdoor Autonomous Driving |
| 2021 | Sep | Hu et al. [75] | - | Domain | Detection | KITTI | Outdoor Autonomous Driving |
| 2021 | Oct | Hahner et al. [76] | - | Domain | Detection | STF | Outdoor Autonomous Driving |
| 2021 | Dec | Nekrasov et al. [77] | Mix3D | Others | Segmentation | ScanNet, S3DIS, SemanticKITTI | Indoor and Outdoor Scene Understanding |
| 2021 | Dec | Yin et al. [78] | - | Multi-modal | Detection | nuScenes | Outdoor Autonomous Driving |
| 2022 | Feb | Xiao et al. [79] | - | Domain | Segmentation | SemanticKITTI, SemanticPOSS | Outdoor Autonomous Driving |
| 2022 | May | Amini et al. [80] | Vista 2.0 | Domain | - | - | Outdoor Autonomous Driving |
| 2022 | Jun | Lehner et al. [81] | 3D-VField | Adversarial Deformation | Detection | KITTI, Waymo, CrashD, SUN RGB-D | Outdoor Autonomous Driving |
| 2022 | Jun | Hahner et al. [82] | - | Domain | Detection | STF | Outdoor Autonomous Driving |
| 2022 | Sep | Zhang et al. [83] | PointCutMix | Mixup | Classification | ModelNet40, ModelNet10, ScanObjectNN, SHREC16, RobustPointSet | 3D Object Understanding |
| 2022 | Oct | Leng et al. [84] | PseudoAugment | Others | Detection | KITTI, Waymo | Outdoor Autonomous Driving |
| 2022 | Oct | Umam et al. [85] | Point MixSwap | Mixup | Classification | ModelNet40, ModelNet10, ScanObjectNN | 3D Object Understanding |
| 2022 | Nov | Lee et al. [86] | SageMix | Mixup | Classification | ModelNet40, ScanObjectNN | 3D Object Understanding |
| 2022 | Nov | Xiao et al. [87] | PolarMix | Others | Segmentation, Detection | SemanticKITTI, nuScenes-lidarseg, SemanticPOSS, nuScenes | Outdoor Autonomous Driving |
| 2022 | Dec | Matuszka et al. [88] | - | Domain | Detection | Argoverse | Outdoor Autonomous Driving |
| 2023 | Jun | Liu et al. [89] | - | Others | Detection | KITTI, Waymo | Outdoor Autonomous Driving |
| 2023 | Jun | Yang et al. [90] | - | Multi-modal | Segmentation | - | Bridge Segmentation |
| 2023 | Jun | Ryu et al. [91] | - | Domain | Segmentation | SemanticKITTI, nuScenes | Outdoor Autonomous Driving |
| 2023 | Oct | Lehner et al. [92] | - | Adversarial Deformation | Detection, Segmentation | KITTI, Waymo, CrashD, SemanticKITTI, nuScenes | Outdoor Autonomous Driving |

Table 2 shows that many specialized augmentation methods for point cloud data started to emerge after 2019. Mixup is the most commonly used specialized augmentation for classification tasks in 3D object understanding. Domain augmentation is predominantly used for detection tasks in outdoor autonomous driving. The utilization of specialized methods to segmentation is comparatively less, usually involving multi-modal, adversarial deformation and domain augmentation techniques as shown in Table 2. It should be noticed that, currently, some adversarial deformation, up-sampling, completion and generation techniques have not been directly applied to point cloud data augmentation, as shown in Table 3. For a comprehensive classification of specialized methods, these potential methods are also included and discussed in this article.

Table 3: Potential specialized point cloud augmentation methods.

| Year | Month | Authors | Name | Augmentation | Application |
|---|---|---|---|---|---|
| 2017 | Oct | Wang et al. [93] | - | Up-sampling | 3D Object Understanding |
| 2017 | Oct | Yang et al. [94] | - | Generation | 3D Object Understanding |
| 2018 | Jun | Groueix et al. [95] | - | Generation, Mixup | 3D Object Understanding |
| 2018 | Jul | Achlioptas et al. [96] | - | Generation, Mixup | 3D Object Understanding |
| 2019 | Apr | Valsesia et al. [97] | - | Generation, Up-sampling | 3D Object Understanding |
| 2019 | May | Li et al. [98] | - | Generation, Mixup | 3D Object Understanding |
| 2019 | Jun | Xiang et al. [99] | - | Adversarial Deformation | 3D Object Understanding |
| 2019 | Jun | Wang et al. [100] | - | Up-sampling | 3D Object Understanding |
| 2019 | Oct | Shu et al. [101] | - | Generation | 3D Object Understanding |
| 2019 | Oct | Yang et al. [102] | - | Generation | 3D Object Understanding |
| 2019 | Oct | Li et al. [103] | - | Up-sampling | 3D Object Understanding |
| 2020 | Apr | Yu et al. [104] | - | Completion | 3D Object Understanding |
| 2020 | Jun | Zhao et al. [105] | - | Adversarial Deformation | 3D Object Understanding |
| 2020 | Dec | Cai et al. [106] | - | Generation | 3D Object Understanding |
| 2020 | Dec | Hamdi et al. [107] | Advpc | Adversarial Deformation | 3D Object Understanding |
| 2021 | May | Yan et al. [108] | - | Completion | Outdoor Autonomous Driving |
| 2021 | Jun | Luo et al. [109] | - | Generation, Mixup | 3D Object Understanding |
| 2022 | Apr | Lyu et al. [110] | - | Completion | 3D Object Understanding |
| 2022 | May | Liu et al. [111] | TauPad | Adversarial Deformation | 3D Object Understanding |
| 2022 | Aug | Wu et al. [112] | - | Adversarial Deformation | 3D Object Understanding |
| 2023 | Nov | Xiong et al. [113] | UltraLiDAR | Generation | Outdoor Autonomous Driving |

### 3.1 Mixup augmentation

Introduced in 2018, Mixup [114] has been widely used for image classification. Its idea is to randomly select two images and mix them in a certain ratio to form a new image. This method was adopted in point cloud classification tasks by some researchers [95, 96], in which two point cloud samples are randomly selected and mixed to generate new training samples. It should be noted that in point cloud processing tasks, mixup is commonly considered to mix two instances and does not typically involve scene-level mixing.

Mixup enables DL models to learn intermediate states between labelled instances. There are various techniques for mixing selected point cloud samples. For example, PointMixup [69] employs an optimal distribution of path functions to find the shortest path between point clouds, generating mixed point cloud samples. However, this method may destroy the structural information of point cloud data [71]. To address this problem, Rigid Subset Mix (RSMix) [71] provides a neighboring function that can extract subsets and perform fusion while preserving the original shape in the point cloud. However, point cloud samples generated by RSMix exhibit noticeable discontinuities at boundaries. To preserve the feature information in point cloud data, SageMix [86] safeguards the structural characteristics of significant local areas within point cloud data during the mixup process, ensuring both continuity and local feature preservation.

PointCutMix [83] establishes point correspondences between two point cloud samples based on the morphing and sampling network [115], performing the mixup operation by exchanging these point correspondences. Figure 7 shows examples of mixed point cloud samples using PointCutMix. FMix [70] provides a novel mixup strategy that generates arbitrarily shaped random masks between two point clouds and swaps points within these masks. However, it performs poorly in ModelNet10 [116], primarily due to the simplistic strategy of exchanging point clouds through mask operations, making it unsuitable for three-dimensional data. Point MixSwap [85], the first end-to-end method to incorporate an attention mechanism into point cloud mixup, uses an attention module to decompose two point cloud samples into several disjoint regions, completing the mixup by exchanging points in these regions. The authors of Point MixSwap tested its effectiveness by applying it to a subset of the original dataset to generate augmented datasets. The test results indicated that this augmentation effectively enhanced the performance of the baseline model in classification tasks. However, the improved performance did not surpass the results obtained when using the whole original dataset.

Certain generative models have the ability to perform mixup operations, often referred to as 'interpolation' by their authors. These models map instances to their latent representations. In the latent space, low-dimensional feature vectors of two samples are weighted and summed to map back to the data space to generate mixup samples. Achlioptas et al. [96] introduced a deep autoencoder (AE) network for point cloud generation, which can be used to perform mixup operations between point cloud data. PC-GAN [98] explores the integration of generative adversarial network (GAN) into point cloud generation, using a smaller network than that of Achlioptas et al. [96] and achieving better generation results. Inspired by non-equilibrium thermodynamic diffusion processes, Luo et al. [109] conceptualized point cloud generation as a back diffusion of noise into a desired shape, performing mixup between two point cloud samples in the latent space.

Mixup demonstrates favorable performance in enhancing model's robustness to datasets with common corruptions [117]. However, mixup tends to significantly destroy semantic information, posing challenges in its adaptation to scenarios beyond classification. Additionally, Sun et al. [117] found through tests that mixup exhibited limited improvement in model performance in common scenarios.

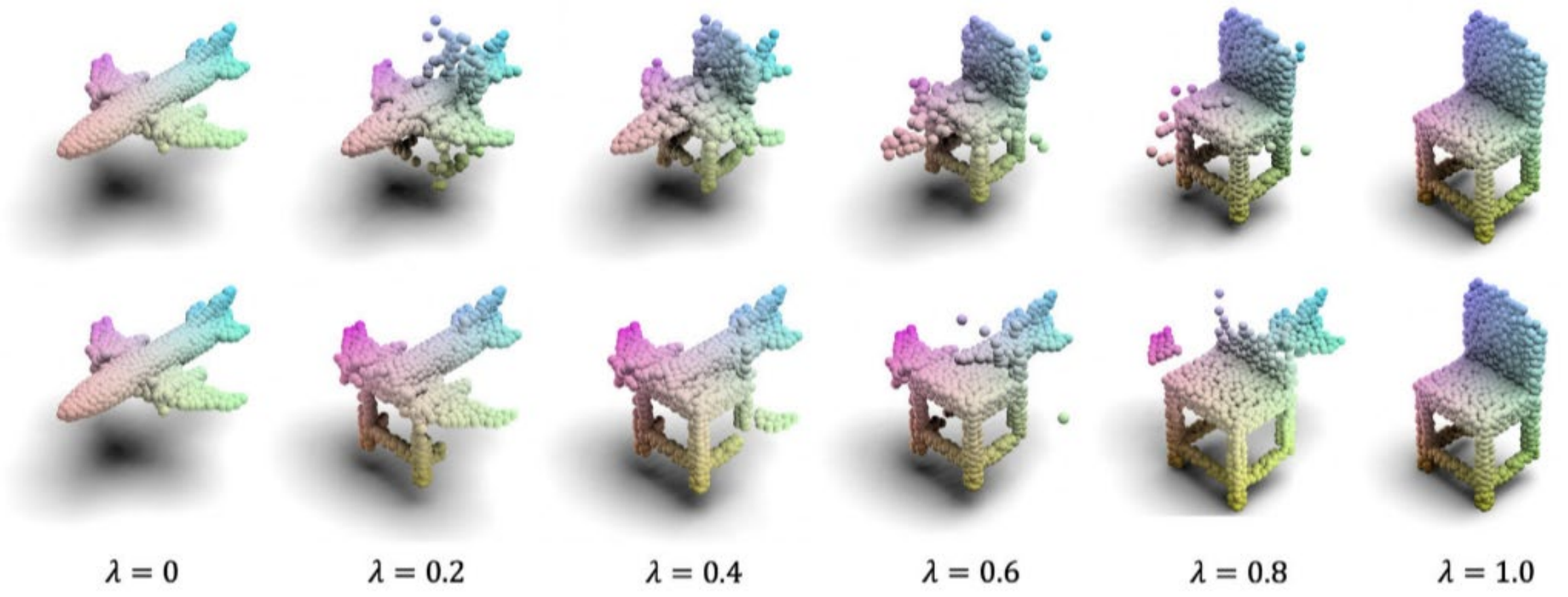

Figure 7: Examples of mixup between different classes at different degrees of mixup (From Zhang et al. [83] with permission of Elsevier).

### 3.2 Domain augmentation

In DL-based point cloud processing tasks, training and testing data may originate from different data domains, such as different environments, locations, time and/or sensors. This type of variations often results in domain gaps characterized by distinct data characteristics, and consequently significantly degrades the performance of point cloud processing models [118]. To address this issue, domain augmentation is often employed to simulate training data from different domains. This augmentation is typically implemented by incorporating samples that mimic the characteristics of various domains into training data.

Changes in the environmental domain, such as severe rainfall and fog, can greatly affect the performance of point cloud processing models [67]. Bijelic et al. [119] collected point cloud data along a distance of over 10,000 kilometers in severe weather. However, these data proved difficult to use in training due to extreme weather conditions. In an effort to improve the resilience of point cloud processing to adverse weather, researchers have simulated point cloud data under harsh environment domains. For example, Hahner et al. [76] used physical modeling to simulate different degrees of fog in their dataset, alleviating the impact of fog on point cloud processing and generating point cloud data under foggy weather for training. Subsequently, Hahner et al. [82] extended their approach to simulate point cloud datasets under snowy weather conditions, as shown in Figure 8. Their method not only modifies beam measurements, but also simulates ground moisture for enhanced realism. Similar research was undertaken by Kilic et al. [74], who simulated point cloud scenes featuring rain, snow and fog.

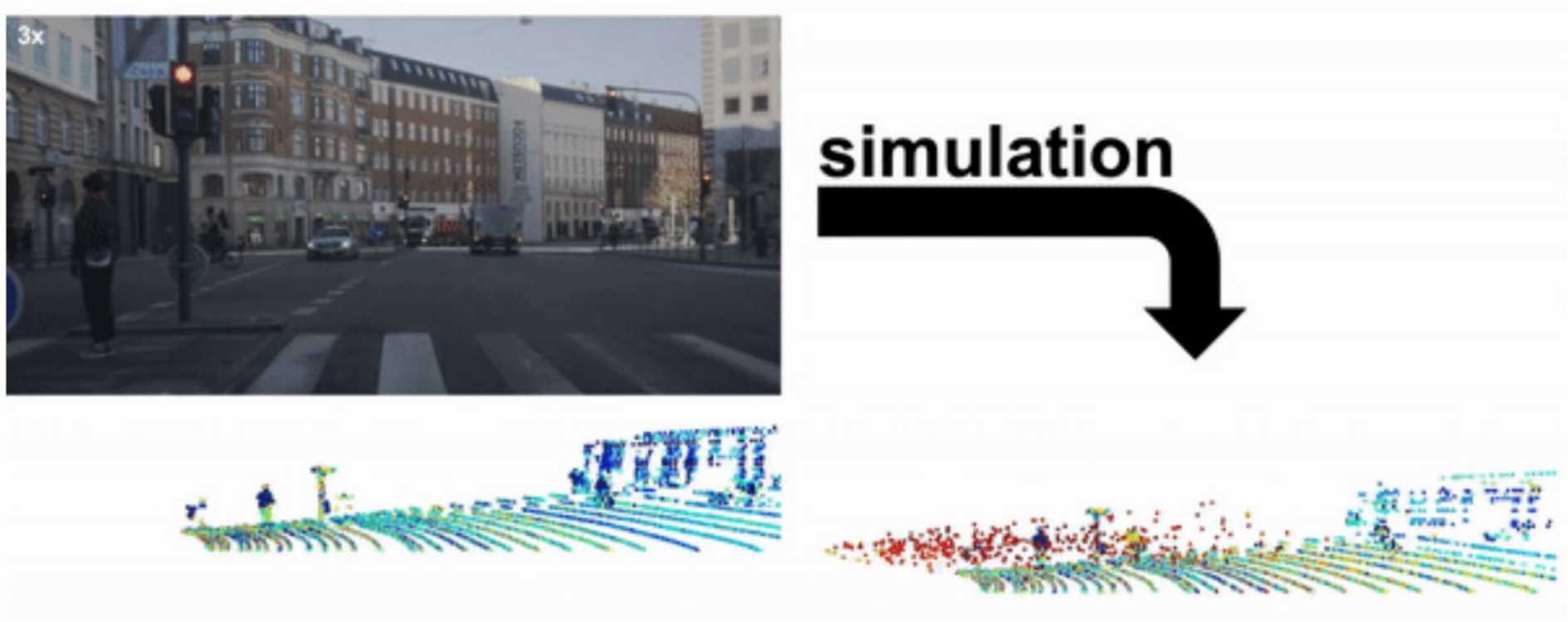

Figure 8: Example of environment domain augmentation. The point cloud data in the lower right corner simulate severe snowfall against the original point cloud data in the lower left corner. The plot at the top shows the RGB image as a reference (From [120] with permission under a Creative Commons License).

The characteristics of point cloud data can vary based on the configuration or model of sensors. Sensor domain augmentation is used to simulate data domains from different sensors. Hu et al. [75] augmented point cloud data by simulating the characteristics of distant point cloud samples according to the attributes of LiDAR sensors. These augmented point cloud data serve as additional training data to improve the DL model's ability to detect instances located farther from sensors. Matuszka et al. [88] extended the two-dimensional zoom augmentation [121] into 3D, simulating the zooming of LiDAR sensors according to LiDAR intrinsics. While keeping the projection unchanged and using virtual sensors, this method simulates the characteristics of point cloud instances at various distances to expand the training set, enabling the model to generalize to distant instances.

To simulate high-resolution point cloud data, Sallab et al. [68] used a sensor model based on CycleGAN [122] to generate simulated dense point cloud data. This method can also generate synthetic data to simulate real-world scenes according to the sensor model. Considering sensor characteristics, Ryu et al. [91] simulated point cloud data from other LiDAR configurations or models to use them as training data,  alleviating the sensor-bias problem.

Synthetic point cloud data exhibit a domain gap when compared to real-world point cloud data, as they are generated by computer engines and tend to be overly idealized. In contrast, real-world data are influenced by many factors such as sensors, lighting and other environmental conditions. To alleviate this domain gap, Xiao et al. [79] proposed a point cloud translator. This translator compares synthetic data with real-world data and reconstructs synthetic point cloud from two aspects: appearance and sparsity. This process aims to refine synthetic point cloud data, making them more closely resemble real-world data. Although this method can narrow the gap between simulated instances and real-world data in autonomous driving scenarios, it fails to address the domain gap related to the background environment of scenes. Data-driven simulations, which involve simulating data based on real-world information, are an effective method for mitigating the domain differences between real-world and simulated data [123]. For instance, Vista 2.0 [80] simulates real-world scenarios of autonomous driving scenes by emulating the structure of onboard LiDAR, generating synthetic point cloud data with real-world domain features.

### 3.3 Adversarial deformation augmentation

Adversarial deformation augmentation refers to using adversarial learning to perturb point cloud data to obtain augmented point cloud data. This augmentation technique enhances the capability of DL models to generalize to variations in the shape of objects in point cloud data, enabling improved adaptation to out-of-domain samples.

Adversarial deformation has been used in numerous studies [72, 81, 92, 112] to augment point cloud data. Attention has been paid to ensuring that deformations are smooth and realistic, accurately representing variations in object shapes in real-world scenarios. For example, 3D-VField [81] uses vector fields to reasonably perturb original point cloud instances, which involves sliding the points of an instance along the ray direction of the sensor view to obtain augmented data. This method can smoothly deform instances' shapes, thereby enhancing the generalization of trained models to instances with varying shapes. It has been proven effective in an autonomous driving dataset involving rare and damaged car instances [81]. Initially designed for detection tasks, this method was subsequently extended to segmentation tasks by Lehner et al. [92]. In this extension, the adversarial loss was redesigned to suit segmentation tasks, with the application of a deformation vector field to instances of specific classes. In addition, this method attempted to adversarially perturb intensity signals to enhance model robustness. Another representative method is Maxup [72], which employs a loss function to encourage the network to smoothly perturb point cloud data without limiting the direction of point deformation.

To promote the use of adversarial deformation augmentation in point cloud processing tasks, Liu et al. [111] proposed TauPad, an easy-to-use tool that applies adversarial deformations on point cloud data for augmentation. However, the utilization of adversarial deformation augmentation is still limited and can be further explored. Various adversarial deformation techniques, not originally intended for data augmentation, generate adversarial samples that can be used as augmentation data. For example, some studies [99, 105, 107] perform adversarial deformations on point cloud data to attack point cloud processing models, which can be readily used for augmentation purposes.

### 3.4 Up-sampling augmentation

Up-sampling is an augmentation technique that algorithmically generates high-resolution (dense) point cloud data from original low-resolution (sparse) point clouds. This method is valuable in mitigating the challenges associated with sparse point cloud data, arising from factors such as the limited measurement resolutions of LiDAR sensors during data acquisition.

Although point cloud up-sampling has not been explicitly designated as an augmentation method by researchers [93, 97, 103], it is considered as a valuable means of augmenting point cloud data, as shown in Figure 9. This method involves two steps: choosing a down-sampling algorithm to down-sample point cloud data in the training set, followed by applying an up-sampling method to the down-sampled data to obtain augmented data. It is essential to note that down-sampling is not always necessary, especially when the initial data are already sparse. Considering semantic rationality, Wang et al. [93] integrated GAN and Recurrent Convolutional Networks (RCN), achieving high-resolution up-sampling results while accounting for context structure and semantic rationality. Leveraging the capability of graph convolution to extract local features from point cloud data, Valsesia et al. [97] successfully integrated graph convolution into GAN for up-sampling tasks, but with relatively high computational complexity. PU-GAN [103], based on GAN, trains a generator network to generate densely distributed point sets, using a discriminator to evaluate and normalize predictions by penalizing deviations from expected generated results. However, this method has limited ability to fill large gaps in point cloud data.

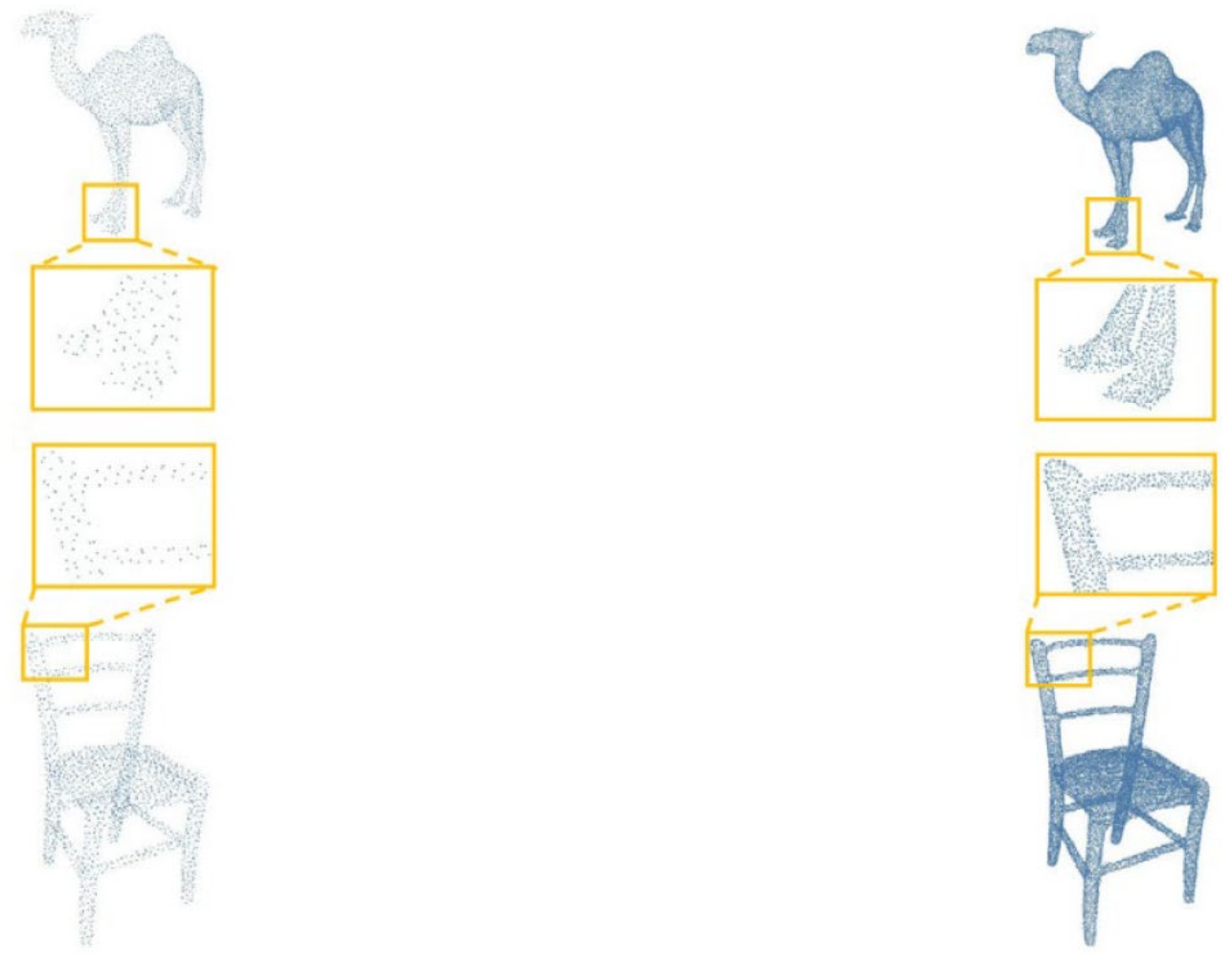

(a) (b)

Figure 9: Examples of augmentation using up-sampling: (a) input data, (b) augmented data (From Hu et al. [124] with permission of Elsevier).

The aforementioned techniques essentially perform data up-sampling, though original point cloud data may initially be down-sampled when necessary. Augmented point cloud data through up-sampling can retain important local semantic information of an instance while at the same time exhibiting slightly different spatial distribution compared to the original data. Although this approach has not been explored by researchers as an augmentation approach, it is deemed to be an innovative means of augmenting point cloud data. The potential of point cloud up-sampling augmentation has been demonstrated in several studies. For instance, Li et al. [103] and He et al. [125] tested their trained models in up-sampled ModelNet test set and achieved improved classification results. Similarly, Wang et al. [126] utilized point cloud up-sampling to enhance the outcomes of point cloud reconstruction, proving that point cloud networks can extract features more effectively from up-sampled point clouds, thereby effectively supporting deep learning tasks involving point clouds [127].

Numerous past point cloud up-sampling methods, trained on synthetic datasets, performed poorly in real-world scans with additional complexity [128]. One influencing factor is its reliance on Chamfer Distance (CD) loss to measure the results, which overlooks detailed structures [129] as well as outlier [130] issues with CD. In downstream tasks that require high-quality representation, the presence of outliers in augmented point cloud data can negatively affect model performance. To mitigate this limitation, Lin et al. [130] calculated the CD loss in hyperbolic space, updating poor matches by adopting a strategy that assigns high weights to matched points with small Euclidean distances. Similarly, Lin et al. [131] improved the CD loss by introducing InfoCD that integrates contrastive learning with the CD loss, learning to disperse matched points to better align the distributions of point clouds. This approach can handle outliers more effectively and improve distribution alignment. Meanwhile, patch-based up-sampling methods, which focus on different levels of detail, prove more capable of recovering the details of reconstructed instances, alleviating performance issues in real-world scans [100]. Wang et al. [100] introduced an adaptive, progressive patch-based up-sampling method, where the network continuously learns the details of patches from previous steps. This method was tested on real-world data and achieved promising upsampling results with local details.

### 3.5 Completion augmentation

Point cloud completion involves prediction of missing parts within point cloud data to generate a complete point cloud [132], as shown in Figure 10. This technique proves beneficial for addressing the difficulties tied to incomplete point cloud data, which result from issues like occlusions and restriction of devices' angles during the data collection process [133].

Although authors [93, 104, 108, 110] focusing on point cloud completion did not explicitly state its use for point cloud data augmentation, it can be tailored for augmentation with the following steps. Initially, part(s) of point cloud data in the training set are deducted to create incomplete data, with the strategy specifically designed according to specific requirements. The subsequent step uses a point cloud completion algorithm to generate complete data serving as augmented data. It is important to note that it is not always necessary to carry out the initial deduction, especially when the initial data are already incomplete. Wang et al.'s method [93] can be used for point cloud completion, but this voxelization-based method may incur information loss [104]. In contrast, Yu et al. [104] proposed an end-to-end network based on point encoder GAN, which has strong generalization ability and prevents additional information loss. Yan et al. [108] completed the point cloud scene through auxiliary information from adjacent frames to enhance point cloud segmentation. However, since information from adjacent frames may be unavailable, the practicality of their method for segmenting point cloud data acquired in many real-world scenarios is limited. For efficient and accurate point cloud completion, Lyu et al. [110] proposed the point diffusion-refinement paradigm, which generates uniform point cloud data by denoising the diffusion probabilistic model and optimizes rough parts in the generated point cloud.

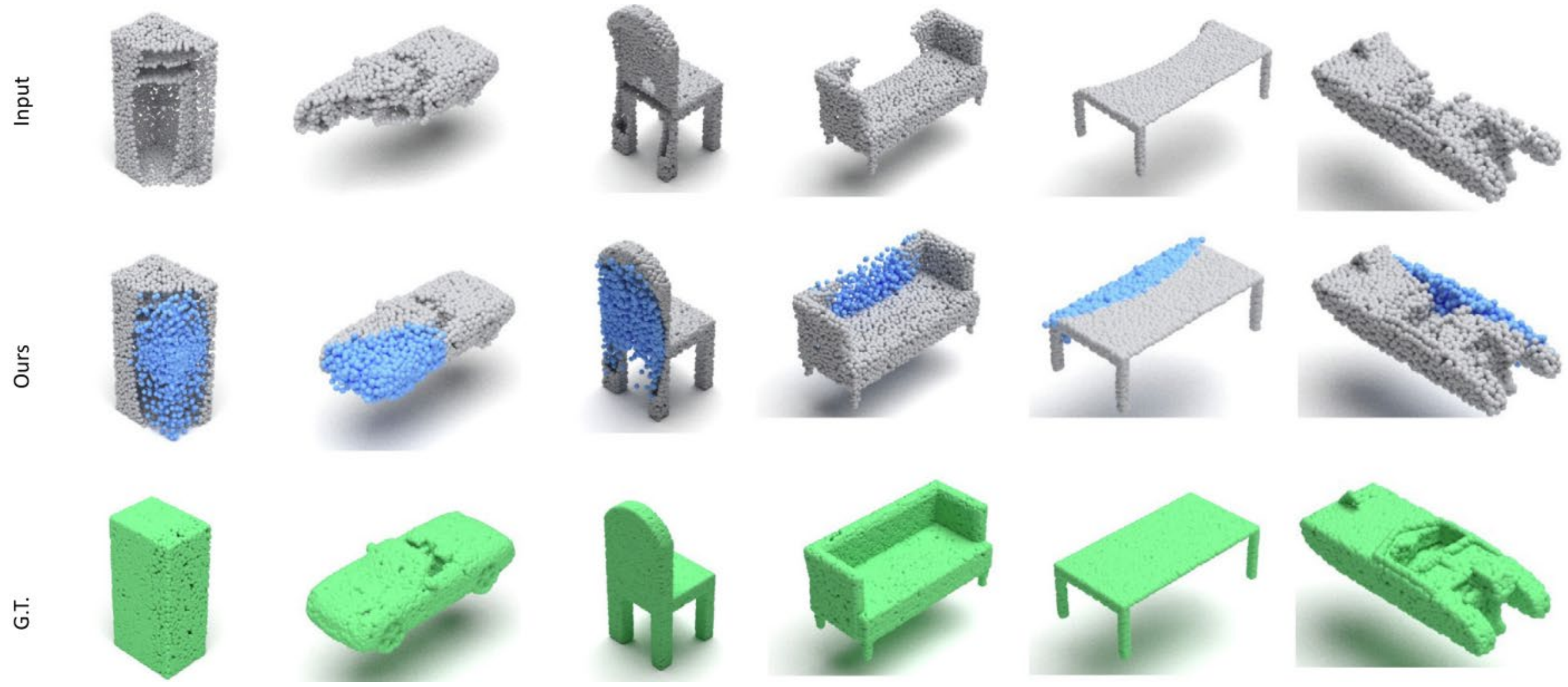

Figure 10: Examples of augmentation using point cloud completion (From Lin et al. [132] with permission of Elsevier).

### 3.6 Generation augmentation

Generation refers to using generative models to train datasets and generate new point cloud instances, aiming to simulate or represent the characteristics of real-world instances in point cloud data, as shown in Figure 11(a). With the advancement of generation models such as GAN and diffusion models, these models have widely been used in various fields [96, 102], but has not been explicitly explored as a method for point cloud data augmentation. However, it is important to recognize that generation augmentation has the capacity to preserve global semantic information while generating new point cloud instances that are significantly different from the original instances, with changes in local characteristics. Therefore, point cloud generation methods can also be used to augment point cloud data.

Generative models can generate diverse synthetic point cloud samples with variations in structure, shape, local details, and more. These synthetic samples can serve the purpose of point cloud data augmentation. The method proposed by Valsesia et al. [97], as discussed in Section 3.4, can also generate diverse point cloud data. Based on a tree-structured graph convolution network, TreeGAN [101] is a GAN designed for unsupervised point cloud data generation. Although its network can better represent features through ancestor information in the network, it comes with high computational complexity. Furthermore, generative models [98, 109] used for mixup augmentation can also be applied for generation augmentation. Figure 11(a) shows examples of synthetically generated point clouds instances. Additionally, scene-level generation has garnered attention. The method proposed by Xiong et al. [113] learns the discrete codebook of real-world autonomous driving scenes through a vector-quantized variational autoencoder [134]. Subsequently, the discrete features mapped by the codebook are decoded to facilitate either conditional or unconditional generation of autonomous driving scenes. Although this method currently cannot simultaneously generate instances such as cars and pedestrians in autonomous driving scenes, its combination with other instance-oriented data augmentation methods is promising.

Point cloud reconstruction represents a special form of point cloud generation. In this operation, a trained generative model is used to reconstruct a partially incomplete or sparse point cloud instance, aiming to obtain dense and accurate point cloud instances that represent the corresponding real-world object. The resulting generated point cloud can be used as augmented data, and visual examples are depicted in Figure 11(b). A representative model in this category is 3D-RecGAN [94], an end-to-end network combining autoencoders and GAN to reconstruct fine and complete 3D structures from voxel grids at various viewing angles. In addition, generative model [95, 96] used for mixup augmentation can also be used for point cloud reconstruction.

Some generative models are capable of generating both synthetic and reconstructed point cloud samples. For instance, the generation model proposed by Luo et al. [109] can generate synthetic point cloud data through a diffusion model, in addition to its capability of point cloud reconstruction. Training GAN often involves high complexity, but to mitigate this issue, Pointflow [102] uses a principled probabilistic framework. This method first learns the distribution of point cloud shapes and then the distribution of points in those shapes, facilitating the generation of synthetic point cloud data or reconstruction of existing point cloud data. Another representative method is ShapeGF [106], a score-based generative model.

Drive-3Daug [135] adopts the concept of generative augmentation to augment images of driving scenes. This method segments images of driving scenes into background and instances, and then uses Neural Radiance Field (NeRF) [136] to reconstruct them into 3D models. New training data are formed by combining reconstructed instances with reconstructed background. Although originally applied to image data, this approach has the potential for extension to point cloud data.

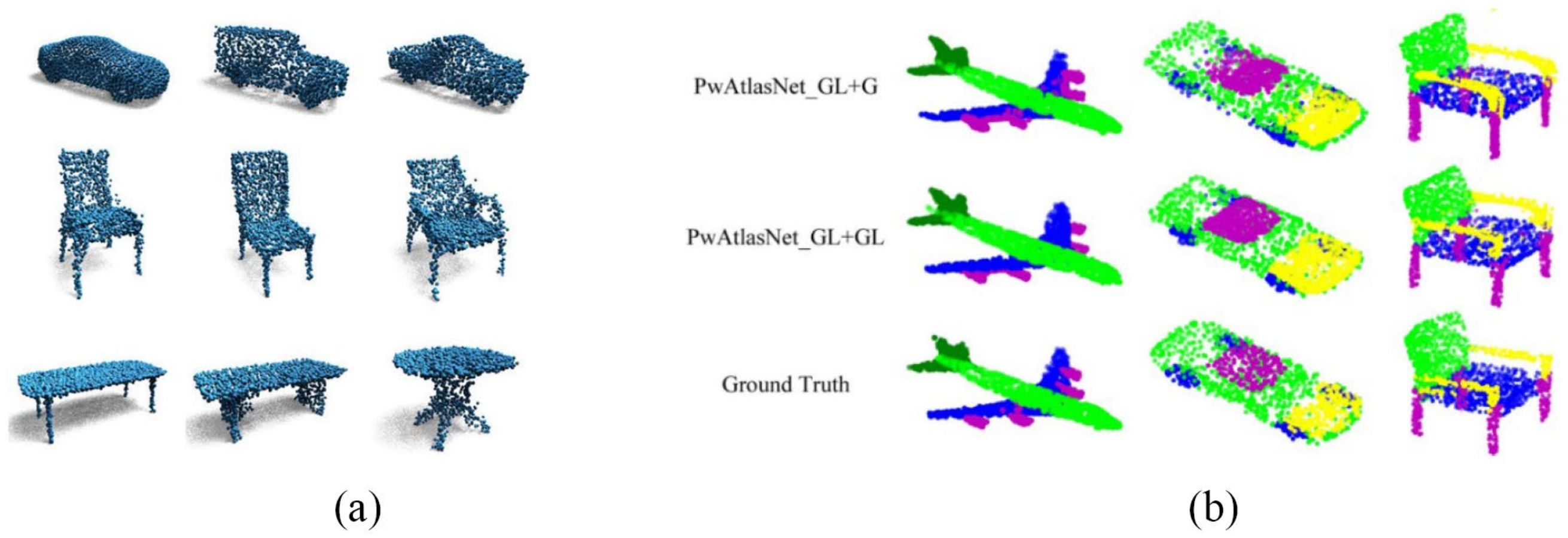

Figure 11: (a) Examples of generated synthetic point clouds instances (From Zamorski et al. [137] with permission of Elsevier). (b) Examples of reconstructed point cloud instances (From Yu et al. [138] with permission of Elsevier).

### 3.7 Multi-modal augmentation

Multi-modal augmentation involves enhancing point cloud data by incorporating data in other modalities. Since images provide complementary features such as semantic information to point cloud data, image modality can be used to augment point cloud data. For example, Yin et al. [78] used an image detection algorithm in the image modality, using the detection results to generate dense point cloud data and augment original sparse point cloud data.

There are cases where enhancing the low resolution of multi-line LiDAR data becomes necessary, typically in autonomous driving scenarios [139]. Typically, this is achieved through super-resolution techniques using high-resolution images. Although both up-sampling and super-resolution can be adopted to augment the density of point clouds, up-sampling does not require image information to obtain denser point clouds. The density augmentation guided by image information in super-resolution may enhance the reconstruction of spatial details in point cloud data, thus improving the overall quality of densified point cloud data. Although previous research [139-141] has demonstrated the effectiveness of super-resolution in increasing the spatial resolution of point cloud data from images, its adaptation to data augmentation is relatively uncommon. This is partly due to more widespread use of cost-effective high-resolution rotary or solid-state LiDAR in autonomous vehicles.

Other data modalities were also explored to augment point cloud data. For example, LiDAR-Aug [73] converts CAD models into point cloud data, randomly placing them in the scene. Yang et al. [90] used synthetic datasets to enhance real-world datasets, employing the geometric partition method to cluster synthetic and real-world point clouds into a point set named superpoints. They then mixed these superpoints in a certain proportion to create a new training set.

However, the implementation of multi-modal augmentation may face certain constraints. In many practical scenarios, acquiring data in the other modality may be challenging or unfeasible. In addition, aligning data from alternative modalities with the point cloud data poses challenges. Furthermore, point cloud augmentation through other modalities may not outperform algorithms specifically designed for multi-modal analysis.

### 3.8 Others

There are also specialized augmentation methods for point cloud data, which do not fall into any of the aforementioned categories due to their uniqueness, such as Mix3D [77] and PolarMix [87]. For example, Mix3D [77] adopts a distinctive fusion approach by blending two scenes to create a new training sample, yielding promising augmentation results. While this method shares a similar concept with mixup augmentation that mixes data at the instance level, it is classified into "Others" to avoid conceptual ambiguity. PolarMix [87] swaps data points within specific local areas in one scene with those in a corresponding local area in another scene for achieving augmentation. Other examples include random noise point generation [38], swap [38, 39, 87] and color [142]), which are seldom used and are not specifically discussed in this review.

In semi-supervised learning, a teacher-student framework is often used where weak augmentation is applied to the teacher network and strong data augmentation is used for the student network [63, 89]. In other words, the student network tends to be more complex or have more data augmentation methods than the teacher network. For example, the teacher network uses only drop, while the student network uses a combination of rotation, flipping, and scaling. Liu et al. [89] designed an effective augmentation method for the student network, involving dividing the point cloud area into rectangles of the same shape using multiple horizontal and vertical lines, shuffling and unshuffling these rectangles in the feature layer. In this way, the student network focuses on weak features, thereby enhancing the feature representation ability. However, this augmentation method may segment complete instances when shuffling, resulting in localization bias.

In the context of semi-supervised learning, where labeled data are used to train a simple model predicting pseudo-labels for unlabeled samples, PseudoAugment [84] uses unlabeled data to augment training data. This method removes low-confidence points in the pseudo-labeled point cloud data generated by semi-supervised learning, subsequently pastes the pseudo-object into the scene, and finally exchanges the pseudo-labeled scene with the labeled scene.

## 4 Discussion and future work

Basic point cloud augmentation methods have been widely used in tasks such as classification, segmentation and/or detection. For example, GT-sampling, when combined with other affine transformation operations, is widely used to enhance scene-level data for detection and segmentation purposes. In contrast, most specialized augmentation methods are usually designed for specific application scenarios. Therefore, specialized augmentation methods are often considered when combinations of basic augmentation operations could not fully fulfill the objective of data augmentation. In this context, specialized methods can also be combined with basic augmentation operations when required and can serve as an important supplement to basic methods in certain application scenarios.

Tables 1 and 2 reveal that the latest advancements in point cloud data augmentation are concentrated in the domains of outdoor autonomous driving and 3D object understanding. When dealing with point cloud data collected in outdoor environments, augmentation strategies frequently involve a combination of affine transformation, dropping, and GT-sampling, or the adoption of domain augmentation, primarily oriented towards enhancing detection tasks. The typical task within the realm of 3D object understanding involves point cloud data classification, wherein synthetic CAD-generated datasets such as ModelNet [116] are commonly utilized due to their flexibility, variability, and cost-effectiveness. In these classification tasks, it is common to apply a combination of affine transformation, dropping, and jittering, or to choose mixup augmentation. However, GT-sampling is seldom employed, given its unsuitability for classification purposes. In the context of indoor point clouds for scene understanding, the choice of basic augmentation techniques is not notably distinct from that for outdoor point clouds, as evidenced in Table 1. This is likely because the choice depends more on actual characteristics of point cloud data regardless of whether they are acquired in indoor or outdoor contexts.

Point cloud data augmentation techniques can be applied to both real-world and synthetic datasets. Based on the surveyed literature, augmentation is commonly implemented in real-world point cloud datasets, such as those representing outdoor autonomous driving scenes captured via vehicle-mounted LiDAR, indoor scenes captured with RGB-D cameras, and objects scanned with RGB-D cameras. Therefore, the discussions in this article primarily center on augmentation in real-world datasets. Nevertheless, many of the augmentation methods identified in this article can also be employed to enhance synthetic point cloud datasets, such as those derived from computer-aided design (CAD) models. Differences in data domains exist between real-world and synthetic datasets [79]. Due to the accessibility of CAD models, a significant body of research, especially in classification tasks [70], relies on ModelNet [116], in addition to ShapeNet [20] that supports some semantic segmentation tasks [37]. With advancements in scene-level generation [113], realistic virtual autonomous driving scenes can now be produced, making such synthetic datasets a promising area for broader application. Although current data augmentation methods do not distinguish between real-world and synthetic datasets, overcoming the differences in data domains is an important goal, as discussed in Section 3.2.

In addition to diversifying training datasets for enhancing model generalization capabilities, as depicted in Figure 12(a), data augmentation also plays an important role in consistency training, as shown in Figure 12(b). Consistency learning, often used in semi-supervised DL, encouraging neural networks to make similar predictions under different input conditions [143], through generating different variants. In image modality, extensive studies [143-145] have demonstrated the effectiveness of consistency training. In weakly supervised learning for point cloud data, a range of studies [146-149] used data augmentation operations to generate variants of point cloud data, facilitating consistency training.

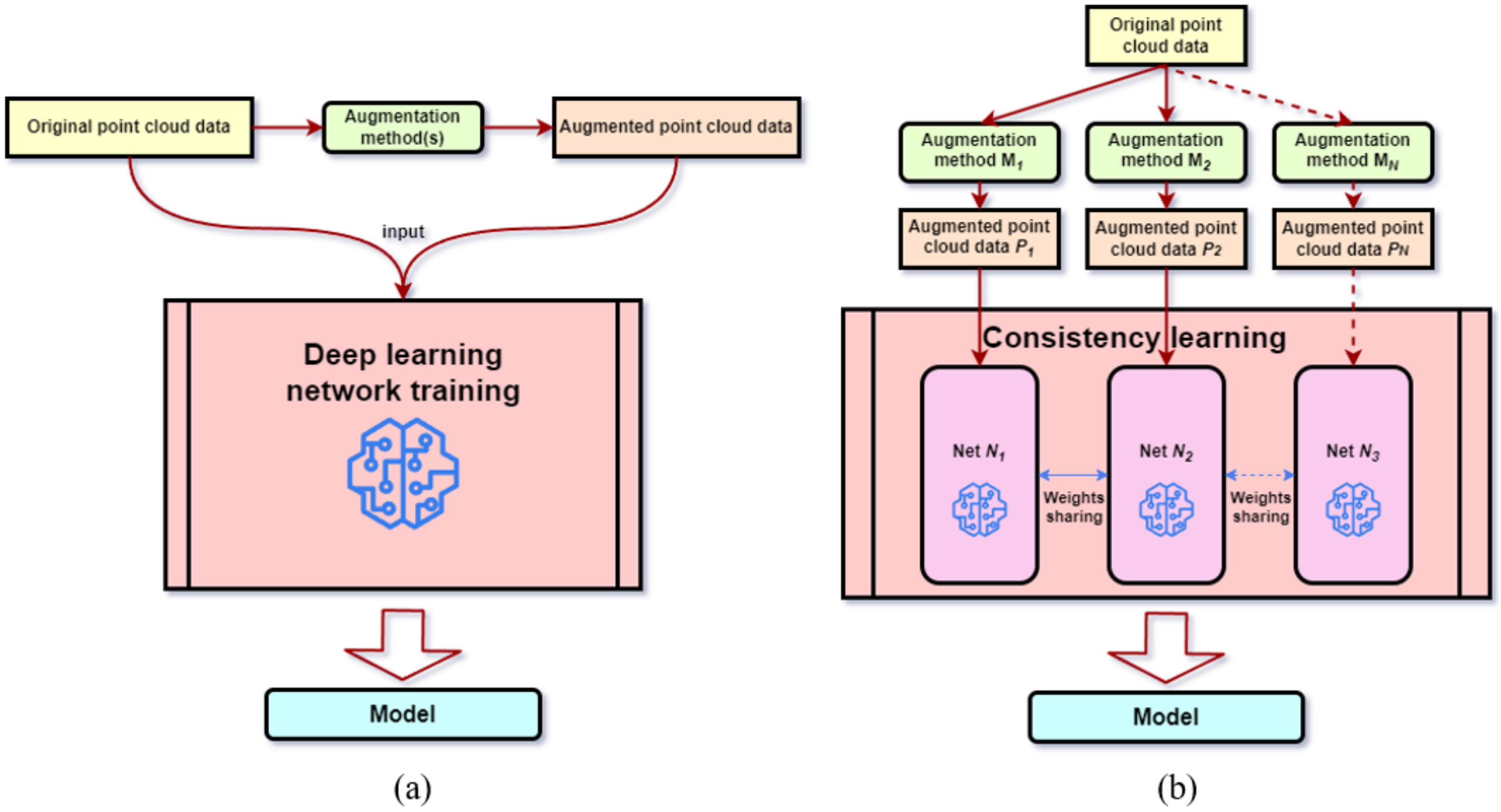

Figure 12: (a) Typical DL training where the original data and the augmented data are sent to DL networks for training to obtain trained models; (b) Consistency learning where the input point cloud data are transformed through various augmentation methods to generate multiple augmented variants, which are then fed into multiple networks for consistency learning to make consistent predictions during training.

We thoroughly examined all surveyed references and identified those containing a quantitative evaluation of model performance before and after data augmentation, commonly detailed within their ablation experiment results. Table 4 presents these identified references along with their reported performance enhancement, as measured by relevant quantitative evaluation metrics. It is evident that integrating data augmentation led to enhanced performance of deep learning models in typical downstream tasks. However, studies conducting such evaluations remain limited. Moreover, due to the lack of consistency in experimental setup (such as datasets and baseline networks), it is still challenging to make fair comparisons across different studies. Existing studies typically involve testing various combinations of augmentation algorithms and DL networks, selecting the combination (i.e., augmentation algorithm and DL network) yielding the highest score for specific point cloud processing tasks. Nevertheless, as another part of our efforts to compare various augmentation methods, Appendix A presents a summary of the quantitative performance of downstream tasks using augmented point cloud data, alongside the adopted augmentation methods in those tasks. This summary includes only those tested with commonly used benchmark datasets. The data presented in Appendix A should be interpreted with caution, as the overall performance is not solely determined by the adopted augmentation methods, but also by the networks utilized for the downstream tasks and the benchmark datasets used for performance evaluation.

Table 4. Reported results on enhanced model performance by point cloud data augmentation. 'GT', 'T', 'R', 'S', 'F', 'D', and 'J' represent GT-sampling, Translation, Rotation, Scaling, Flipping, Drop, and Jittering, respectively. The term 'Without/With Augmentation' indicates performances achieved using original data and augmented data, respectively, for training. The term 'Enhancement Percentage' represents the percentage of performance improvement attributed to data augmentation. The performance metrics used for classification, segmentation, and detection tasks are accuracy, mIoU, and mAP, respectively, with all values represented in percentage (%).

| Augmentation | | Authors | Name | Baseline | Task | Dataset | Without/With Augmentation | Enhancement Percentage |
|---|---|---|---|---|---|---|---|---|
| Basic | D | He et al. [15] | - | - | Classification | ModelNet40 | 92.30/93.00 | 0.70 |
| | T+R+S+J | Li et al. [61] | PointAugment | DGCNN | Classification | ModelNet40 | 92.20/93.40 | 1.20 |
| | T+R+S+D+ J Patch | Sheshappanavar et al. [41] | PatchAugment | DGCNN | Classification | ModelNet40 | 92.20/93.10 | 0.90 |
| | T+S+F+D | Hasecke et al. [35] | - | Cylinder3D | Segmentation | SemanticKITTI | 63.90/65.40 | 1.50 |
| | R+S+J | Lai et al. [37] | - | - | Segmentation | S3DIS | 66.10/72.00 | 5.90 |
| | T+R+S+D+ J Auto+ Patch | Kim et al. [40] | PointWOLF | DGCNN | Segmentation | ShapeNetPart | 84.80/85.20 | 0.40 |
| | GT+R+S+F | Xiao et al. [32] | - | - | Detection | KITTI Validation Car Moderate | 78.63/79.46 | 0.83 |
| | GT+T+R+S +F | Zhu et al. [33] | GT-AUG | - | Detection | nuScenes Validation | 35.68/42.64 | 6.96 |
| | GT+T+R+S +F | Hu et al. [31] | CA-AUG | PVRCNN | Detection | KITTI Validation Moderate | 70.75/72.96 | 2.21 |
| | T+R+S+F+ D Auto | Cheng et al. [62] | PPBA | StarNet | Detection | KITTI Test Car Moderate | 73.99/77.65 | 3.66 |
| | T+R+S+F+D Auto | Leng et al. [64] | LidarAugment | SWFormer_3f | Detection | WOD L2 Test | 73.40/74.80 | 1.40 |
| | T+R+S+F+ D Patch | Choi et al. [38] | - | PVRCNN | Detection | KITTI Validation Car Moderate | 71.21/72.21 | 1.00 |
| | T+R+S+F+ D Patch | Zheng et al. [39] | - | - | Detection | KITTI Validation Car Moderate | 83.22/83.70 | 0.48 |
| Specialized | Mixup | Chen et al. [69] | PointMixup | PointNet++ | Classification | ModelNet40 | 91.90/92.70 | 0.80 |
| | Mixup | Harris et al. [70] | Fmix | PointNet | Classification | ModelNet10 | 89.10/89.57 | 0.47 |
| | Mixup | Lee et al. [71] | RSMix | DGCNN | Classification | ModelNet40 | 92.80/93.50 | 0.70 |
| | Mixup | Zhang et al. [83] | PointCutMix | DGCNN | Classification | ModelNet40 | 92.20/93.20 | 1.00 |
| | Mixup | Umam et al. [85] | Point MixSwap | DGCNN | Classification | ModelNet40 | 92.70/93.50 | 0.80 |
| | Mixup | Lee et al. [86] | SageMix | DGCNN | Classification | ModelNet40 | 92.90/93.60 | 0.70 |
| | Domain | Xiao et al. [79] | - | - | Segmentation | SemanticKITTI | 60.30/64.70 | 4.40 |
| | Others | Nekrasov et al. [77] | Mix3D | KPConv | Segmentation | ScanNet Validation | 69.30/70.30 | 1.00 |
| | Others | Xiao et al. [87] | PolarMix | SPVCNN | Segmentation | SemanticKITTI | 58.00/66.20 | 8.20 |
| | Domain | Hu et al. [75] | - | PVRCNN | Detection | KITTI Validation Car Moderate | 84.50/84.79 | 0.29 |
| | Multi-modal | Fang et al. [73] | LiDAR-Aug | PVRCNN | Detection | KITTI Validation Car Moderate | 78.83/84.23 | 5.40 |

Based on our survey and understanding, the subsequent bullet points highlight a range of existing issues or gaps associated with current point cloud data augmentation, followed by suggested potential directions for future research.

(1) Researchers have not adequately investigated adversarial deformation, up-sampling, completion and generation for performing point cloud data augmentation. Given the advancements in GAN and diffusion models, these can be used for generating realistic and diverse point cloud instances. Future studies should evaluate these approaches on benchmark datasets specific to particular point cloud processing tasks, to assess their effectiveness as augmentation techniques.

(2) Currently, there are few studies dedicated to evaluate the performance of point cloud data augmentation methods using consistent baseline networks and datasets for various point cloud processing tasks. Such evaluations will enhance our understanding of the performance of different augmentation methods. Therefore, future research efforts may focus on establishing novel approaches, metrics and/or datasets to evaluate the effectiveness of point cloud data augmentation methods and their impact on the performance of DL models.

(3) Certain specialized augmentation methods can be computationally expensive when being applied to large-scale point cloud datasets. Future work can focus on developing efficient algorithms with a trade-off between computational cost and augmentation effectiveness. Additionally, some specialized point cloud augmentation methods are relatively complex to be replicated. It is suggested to develop plug-and-play methods, facilitating their widespread adoption. For example, creating a data augmentation library for specialized methods within the PyTorch framework would be beneficial.

(4) There is a lack of a universally accepted combinations of basic augmentation operations for point cloud data augmentation. Therefore, future work needs to establish a standard protocol for selecting augmentation operations for different application domains, tasks and/or datasets, without sacrificing augmentation effectiveness.

(5) Multiple point cloud variants generated through augmentation can affect the effectiveness of consistency learning [149]. Currently, to our knowledge, only basic augmentation methods are used in consistency learning, as seen in PointMatch [148] where basic methods such as scaling, rotation, flipping and jittering were used. Exploring specialized point cloud augmentation methods, such as adversarial deformation and generation augmentation, presents an interesting approach for improving the effectiveness of consistency learning, which is considered a valuable future research direction.

(6) Currently, limited research has attempted the combination of basic and specialized point cloud augmentation methods. Such a combination has the potential to further increase the versatility of data augmentation, deserving future investigation. For example, the fusion of GT-sampling and generation methods can effectively extend generation-based augmentation to scene-level data, such as in autonomous driving scenes. Specifically, this involves sampling instances of cars and pedestrians through GT-sampling, while new point cloud instances are generated using generation augmentation and placed at reasonable locations in the dataset.

(7) Augmentation needs to realistically simulate variations in point cloud data, such as changes in object size, locations, orientation, appearance and environment, to ensure that simulated data align with real-world situations and remains semantically correct. Future research could look into standardizing ranges for various augmentations, tailored to specific application scenarios.

(8) Certain applications such as object detection may involve dynamic objects in the scene. Point clouds captured in dynamic environments may require specialized augmentation strategies that consider the temporal variations of objects. For example, a specific trajectory of moving objects may be designed, which can be achieved by a set of combined augmentation operations such as translation, rotation, and dropping (i.e., simulated occlusion caused by moving objects or changes in point cloud data density as objects moving to or away from sensors).

(9) Transformer networks have achieved strong performance in segmentation and classification tasks even with a simple combination of basic operations. It would be interesting to explore how augmentations perform when integrated with state-of-the-art Transformers as backbone networks.

## 5 Conclusion

The development of DL methods for various point cloud processing tasks has rapidly increased the demand for larger and more diverse training datasets to enhance the performance of DL models. In this regard, data augmentation is widely adopted, especially in cases where there is a limited quantity of annotated point cloud data. This article provides the first comprehensive survey of diverse point cloud data augmentation methods structured within a proposed taxonomy framework, and discusses their strengths and limitations. Based on our survey, a set of potential directions for future research is proposed.

Our survey reveals that basic augmentation remains the preference for most point cloud perception tasks due to simplicity and adaptability through various combinations. These characteristics underpin their widespread utilization in various tasks. While certain basic augmentation techniques, such as rotation and drop, are occasionally used alone, a combination of several basic operations is frequently adopted. GT-sampling, when combined with other affine transformation operations, are commonly used for detection tasks, occasionally for segmentation tasks, but rarely for classification tasks. The choice of individual basic operations for combination is often based on extensive trial experiments or practical experience. Nevertheless, efforts have been made to develop algorithms for automatically determining the optimal combination of basic operations and their associated parameters.

Typically aimed at addressing specific challenges, specialized augmentation employs finer strategies to augment point cloud data. Mixup stands out as the most widely employed technique for enhancing models' performance in classification tasks in the realm of 3D object understanding. On the other hand, domain augmentation takes precedence in detection tasks, particularly in the context of outdoor autonomous driving scenarios. Despite specialized methods have attracted increasing attention from researchers, they have not shaken the prevalent adoption of basic methods in many point cloud processing tasks, primarily due to their task-specific nature and the greater complexity of their implementation. Therefore, the development of specialized methods should be oriented towards being plug-and-play and adaptable to a wide range of application scenarios. Most specialized augmentation methods are designed to meet special needs in a particular point cloud processing task, and can be considered as supplementary techniques to basic methods. Therefore, integrating specialized augmentation with basic augmentation (e.g., generation and GT-sampling) represents a more comprehensive means of data augmentation. In addition, adversarial deformation, up-sampling, completion and generation have not yet been widely adopted for data augmentation in existing studies, but are deemed to have great potential in augmenting point cloud data.

Our survey confirms that incorporating point cloud data augmentation can effectively improve the performance of deep learning models in various downstream tasks, as indicated in Table 4. Although Appendix A presents a summary of the quantitative performance of downstream tasks using augmented point cloud data from existing studies, our survey does not offer a quantitative comparison of the performance of various augmentation methods across those studies. This task is currently challenging because different studies typically involve testing different augmentation methods with different DL networks for downstream tasks on various benchmark datasets, resulting in a lack of consistent benchmarks (e.g., benchmark datasets and baseline networks) for fair comparison. Therefore, novel approaches, metrics and/or datasets need to be established to evaluate the effectiveness of point cloud data augmentation methods and their impact on the performance of DL models. Furthermore, for the same point cloud processing task, the choice of augmentation methods varies in existing studies. It would be beneficial to establish a standard protocol for selecting augmentation methods, while taking into account the characteristics of datasets and application tasks.

## Declaration of competing interest

The authors declare that they have no known competing financial interests or personal relationships that could have appeared to influence the work reported in this paper.

## Acknowledgments

This work was supported in part by the Xi'an Jiaotong-Liverpool University Research Enhancement Fund under Grant REF-21-01-003, and in part by the Xi'an Jiaotong-Liverpool University Postgraduate Research Scholarship under Grant FOS2210JJ03.

**Qinfeng Zhu** received the degree of Master of Research in Computer Science from the University of Liverpool, Liverpool, U.K., in 2023, where he is currently working towards the Ph.D. degree in Computer Science. His research interests mainly lie on deep learning, especially in multi-modal information fusion, 3D computer vision, and data augmentation.

**Lei Fan** received the Ph.D. degree from the University of Southampton, Southampton, U.K., in 2014. He is currently an Assistant Professor with the Department of Civil Engineering, Xi'an Jiaotong-Liverpool University, Suzhou, China. His main research interests include lidar and photogrammetry techniques, point cloud, machine learning, deformation monitoring, semantic segmentations, monitoring of civil engineering structures, and geohazards.

**Ningxin Weng** received the M.Sc. degree in Hydrology and Water Resources Management from the Imperial College London, London, U.K., in 2020. She is currently working toward the Ph.D. degree in civil engineering in the University of Liverpool, Liverpool, U.K. Her research interests include deep learning interpretability and augmentation of remote sensing data.

# Appendix A. Reported Performance in Downstream Tasks

As a component of our work to compare different augmentation methods, Appendix A provides a tabular summary of the quantitative performance of downstream tasks (detection, segmentation and classification) utilizing augmented point cloud data, along with the point cloud data augmentation methods employed in those tasks. This summary comprises only those tasks tested with commonly used benchmark datasets (i.e., KITTI [1], nuScence [2] and Waymo [3] for detection tasks in Tables 1-4, SemanticKITTI [4] and ShapeNet [5] for segmentation tasks in Table 5, and ModelNet10, ModelNet40 [6] and ScanObjectNN [7] for classification tasks in Table 6).

As discussed in our article, quantifying the impact of augmentation methods on the overall performance of downstream tasks presents a challenge. Current research typically involves experimenting with various combinations of augmentation algorithms and DL networks to determine the combination (i.e., augmentation algorithm and DL network for a specific downstream task) that yields the highest score. Therefore, it is advisable to interpret the data presented in Appendix A with caution, as the overall performance is influenced not only by the augmentation methods employed but also by the networks utilized for the downstream tasks and the benchmark datasets used for performance evaluation.

Table 1. Comparative results of representative methods for the detection task on the KITTI 3D test benchmark.

Notes applicable to Tables 1, 2 and 3: 'E', 'M', and 'H' represent the easy, moderate, and hard categories of objects in this benchmark dataset, respectively, as evaluated using Average Precision (AP) metrics. The Intersection over Union (IoU) threshold for 3D bounding boxes is 0.7 for Cars and 0.5 for Pedestrians and Cyclists. The '%' after the value is omitted. '-' means that the result is not available. The values in italics come from replicated tests by others. 'GT', 'T', 'R', 'S', 'F', 'D', 'J' represent GT-sampling, Translation, Rotation, Scaling, Flipping, Drop, Jittering, respectively.

| Method | | | Baseline | Speed (fps) | Cars | | | Pedestrians | | | Cyclists | | | 3D mAP |
|---|---|---|---|---|---|---|---|---|---|---|---|---|---|---|
| | | | | | E | M | H | E | M | H | E | M | H | Moderate |
| Basic | SA-SSD [8] | GT+T+R+S+F | - | 25.00 | 88.75 | 79.79 | 74.16 | - | - | - | - | - | - | - |
| | PointRCNN [9] | GT+T+R+S+F | - | *10.00* | 85.94 | 75.76 | 68.32 | 49.43 | 41.78 | 38.63 | 73.93 | 59.60 | 53.59 | 59.05 |
| | SECOND [10] | GT+T+R+S | - | 20.00 | 83.13 | 73.66 | 66.2 | 51.07 | 42.56 | 37.29 | 70.51 | 53.85 | 46.90 | 56.69 |
| | PointPillars [11] | GT+T+R+S+F | - | 62.00 | 79.05 | 74.99 | 68.3 | 52.08 | 43.53 | 41.49 | 75.78 | 59.07 | 52.92 | 59.20 |
| | 3D-VDNet [12] | GT+R+S+F | - | 38.00 | 87.13 | 78.05 | 72.9 | - | - | - | - | - | - | - |
| | STD [13] | GT+T+R+S+F | - | 10.00 | 86.61 | 77.63 | 76.06 | **53.08** | **44.24** | **41.97** | 78.89 | 62.53 | 55.77 | 61.47 |
| | PV-RCNN [14] | GT+R+S+F | - | *12.50* | 90.25 | 81.43 | 76.82 | - | - | - | 78.60 | 63.71 | **57.65** | - |
| | SE-SSD [15] | Patch-based (T+R+S+F+D) | - | 32.00 | **91.49** | **82.54** | 77.15 | - | - | - | - | - | - | - |
| | PPBA [16] | Auto optimization (T+R+S+F+D) | StarNet [17] | - | 84.16 | 77.65 | 71.21 | 52.65 | 44.08 | 41.54 | 79.42 | 61.99 | 55.34 | 61.24 |
| | Point-GNN [18] | T+R+F+J | - | - | 88.33 | 79.47 | 72.29 | 51.92 | 43.77 | 40.14 | 78.60 | 63.48 | 57.08 | 62.24 |
| Specialized | 3D-VField [19, 20] | Adversarial deformation | Part-A^2 [21] | *11.15* | 89.65 | 79.26 | 78.62 | - | - | - | - | - | - | - |

Table 2. Comparative results of representative methods for the detection task on the KITTI 3D validation benchmark.

| | Method | | Baseline | Speed (fps) | Cars | | | Pedestrians | | | Cyclists | | | 3D mAP Moderate |
|---|---|---|---|---|---|---|---|---|---|---|---|---|---|---|
| | | | | | E | M | H | E | M | H | E | M | H | |
| Basic | CA-AUG [22] | GT+T+R+S+F | PV-RCNN [14] | - | **92.25** | **84.93** | **82.64** | 66.00 | 59.77 | 55.41 | **92.58** | **74.19** | **69.76** | 72.96 |
| | PointRCNN [9] | GT+T+R+S+F | - | *10.00* | 88.88 | 78.63 | 77.38 | *62.16* | *58.00* | *50.53* | *91.72* | *72.47* | *68.18* | *69.70* |
| | PV-RCNN [14] | GT+R+S+F | - | *12.50* | *91.80* | *84.50* | *82.42* | *64.77* | *57.20* | *52.43* | *91.20* | *72.20* | *68.62* | *71.30* |
| | SECOND [10] | GT+T+R+S | - | 20.00 | *90.77* | *81.95* | *78.91* | *56.88* | *52.96* | *48.22* | *82.40* | *64.09* | *59.69* | *66.33* |
| | 3D-VDNet [12] | GT+R+S+F | - | 38.00 | 90.15 | 81.66 | 78.97 | - | - | - | - | - | - | - |
| | STD [13] | GT+T+R+S+F | - | 10.00 | 89.70 | 79.80 | 79.30 | 73.90 | **66.60** | **62.90** | 88.50 | 72.80 | 67.90 | **73.07** |
| | PA-AUG [23] | Patch-based (T+R+S+F+D) | PV-RCNN [14] | - | 89.38 | 80.90 | 78.95 | 67.57 | 60.61 | 56.58 | 86.56 | 72.21 | 68.01 | 71.24 |
| Specialized | Pattern-Aware GT [24] | Domain augmentation | PV-RCNN [14] | - | 92.13 | 84.79 | 82.56 | 65.99 | 58.57 | 53.66 | 90.38 | 72.03 | 67.96 | 71.80 |
| | LiDAR-Aug [25] | Multi-modal | PV-RCNN [14] | - | 90.18 | 84.23 | 78.95 | 65.05 | 58.90 | 55.52 | - | - | - | - |

Table 3 Comparative results of representative methods for the detection task on the KITTI BEV test benchmark.

| | Method | | Baseline | Speed (fps) | Cars | | | Pedestrians | | | Cyclists | | | BEV mAP Moderate |
|---|---|---|---|---|---|---|---|---|---|---|---|---|---|---|
| | | | | | E | M | H | E | M | H | E | M | H | |
| Basic | SA-SSD [8] | GT+T+R+S+F | - | 25.00 | 95.03 | 91.03 | 85.96 | - | - | - | - | - | - | - |
| | SECOND [10] | GT+T+R+S | - | 20.00 | 88.07 | 79.37 | 77.95 | 55.10 | 46.27 | 44.76 | 73.67 | 56.04 | 48.78 | 60.56 |
| | PointPillars [11] | GT+T+R+S+F | - | 62.00 | 88.35 | 86.1 | 79.83 | 58.66 | 50.23 | **47.19** | 79.14 | 62.25 | 56.00 | 66.19 |
| | STD [13] | GT+T+R+S+F | - | 10.00 | 89.66 | 87.76 | **86.89** | **60.99** | **51.39** | 45.89 | 81.04 | 65.32 | 57.85 | 68.16 |
| | PV-RCNN [14] | GT+R+S+F | - | *12.50* | 94.98 | 90.65 | 86.14 | - | - | - | 82.49 | 68.89 | 62.41 | - |
| | 3D-VDNet [12] | GT+R+S+F | - | 38.00 | 91.72 | 88.15 | 84.65 | - | - | - | - | - | - | - |
| | SE-SSD [15] | Patch-based (T+R+S+F+D) | - | 32.00 | **95.68** | **91.84** | 86.72 | - | - | - | - | - | - | - |
| | Point-GNN [18] | T+R+F+J | - | - | 93.11 | 89.17 | 83.9 | 55.36 | 47.07 | 44.61 | 81.17 | 67.28 | 59.67 | 67.84 |
| | PIXOR [26] | R+F | - | 10.75 | 81.70 | 77.05 | 72.95 | - | - | - | - | - | - | - |

Table 4. Comparative results of representative methods for the detection task on the nuScenes and Waymo benchmarks. On the nuScenes benchmark, mean Average Precision (mAP) and nuScenes detection score (NDS) are used for evaluation metrics. On the Waymo validation benchmark, mAP and mean Average Precision weighted by Heading (mAPH) are used. The IoU threshold for 3D bounding boxes is 0.7 for Vehicle and 0.5 for Pedestrians and Cyclists. The '%' after the value is omitted. '-' means that the result is not available. The values in italics come from replicated tests by others. 'GT', 'T', 'R', 'S', 'F', 'D' represent GT-sampling, Translation, Rotation, Scaling, Flipping, Drop, respectively.

| | Method | | Baseline | nuScenes | | | | Waymo v1.2 Validation | | | | | |
|---|---|---|---|---|---|---|---|---|---|---|---|---|---|
| | | | | Validation | | Test | | Vehicle | | Pedestrian | | Cyclist | |
| | | | | mAP | NDS | mAP | NDS | L1 mAP\ mAPH | L2 mAP\ mAPH | L1 mAP\ mAPH | L2 mAP\ mAPH | L1 mAP\ mAPH | L2 mAP\ mAPH |
| Basic | MEGVII [27] | GT+T+R+S+F | - | 52.8 | **63.3** | - | - | - | - | - | - | - | - |
| | PointAugmenting [28] | GT+T+R+S+F | - | - | - | **66.8** | **71.0** | 67.41/- | 62.70/- | 75.42/- | 70.55/- | **76.29/-** | **74.41/-** |
| | PointPillars [11] | GT+T+R+S+F | - | *31.5* | - | *30.5* | *45.3* | *63.30/62.70* | *55.20/54.70* | *68.90/56.60* | *60.40/49.10* | - | - |
| | Hu et al. [29] | GT+T+R+F | - | 35.4 | - | 35.0 | - | - | - | - | - | - | - |
| | SECOND [10] | GT+T+R+S | | *48.8* | *58.6* | *31.6* | *46.8* | *72.27/71.69* | *63.85/63.33* | *68.70/58.18* | *60.72/51.31* | *60.62/59.28* | *58.34/57.05* |
| | PV-RCNN [14] | GT+R+S+F | - | *46.7* | *53.4* | - | - | 77.51/76.89 | 68.98/68.41 | 75.01/65.65 | 66.04/57.61 | 67.81/66.35 | 65.39/63.98 |
| | LidarAugment [30] | Auto optimization (T+R+S+F+D) | SWFormer _3f [31] | - | - | - | - | 80.90/**80.40** | 72.80/72.40 | **84.40/80.70** | 76.80/73.20 | - | - |
| | 3D Auto Labeling [32] | T+R+S+F | - | - | - | - | - | **84.50**/- | - | 82.88/- | - | - | - |
| | AFDetV2 [33] | T+R+S | - | - | - | 62.4 | 68.5 | - | **77.64/77.14** | - | **80.19/74.62** | - | 73.72/**72.24** |
| Specialized | Yin et al. [34] | Multi-modal | - | - | - | 46.4 | 58.1 | - | - | - | - | - | - |
| | PolarMix [35] | Others | CenterNet [36] | **55.4** | 61.1 | - | - | - | - | - | - | - | - |

Table 5. Comparative results of representative methods for the segmentation task on the ShapeNetPart and SemanticKITTI benchmark. Overall Accuracy (OA) and mIoU are used for evaluation metrics. The '%' after the value is omitted. '-' means that the result is not available. 'GT', 'T', 'R', 'S', 'F', 'D', 'J' represent GT-sampling, Translation, Rotation, Scaling, Flipping, Drop, Jittering, respectively.

| Method | | | Baseline | ShapeNetPart (mIoU) | | SemanticKITTI | | |
|---|---|---|---|---|---|---|---|---|
| | | | | **Cat (mIoU)** | **Ins (mIoU)** | **Speed (fps)** | **Test (mIoU)** | **Validation (mIoU)** |
| Basic | PolarNet [37] | R | - | - | - | 16.20 | 54.3 | - |
| | RangeFormer [38] | GT+T+R+S+F+D+J | - | - | - | - | **73.3** | - |
| | Hasecke et al. [39] | GT+T+S+F+D | Cylinder3D [40] | - | - | - | 65.4 | **67.9** |
| | Panoptic-PolarNet [41] | GT+T+R | - | - | - | 11.62 | 59.5 | 64.5 |
| | PointWOLF [42] | Patch-based (T+R+S+D+J) | DGCNN [43] | - | 85.2 | - | - | - |
| | Stratified Transformer [44] | R+S+J | - | **85.1** | **86.6** | - | - | - |
| | PointNet [45] | R+J | - | 80.4 | 83.7 | - | *14.6* | - |
| | PointNet++ [46] | T+R | - | *81.9* | *85.1* | | *20.1* | - |
| | MaskPoint [47] | T+S | - | 84.4 | 86.0 | - | - | - |
| | FPS-Net [48] | R+F | - | - | - | - | 57.1 | - |
| | CrossPoint [49] | T+R+S+F+D+J | - | - | 85.5 | - | - | - |
| Specialized | Mix3D [50] | Others | Rigid KPConv [51] | - | - | - | 63.6 | 66.6 |
| | PolarMix [35] | GT+R+S | SPVCNN [52] | - | - | - | - | 66.2 |

Table 6. Comparative results of representative methods for the classification task on the ModelNet and ScanObjectNN benchmark. OA is used for evaluation metrics. The '%' after the value is omitted. '-' means that the result is not available. The values in italics come from replicated tests by others. 'T', 'R', 'S', 'F', 'D', 'J' represent Translation, Rotation, Scaling, Flipping, Drop, Jittering, respectively.

| Method | | | Baseline | ModelNet | | ScanObjectNN | |
|---|---|---|---|---|---|---|---|
| | | | | **ModelNet40(OA)** | **ModelNet10(OA)** | **OBJ_ONLY (OA)** | **PB_T50_RS (OA)** |
| Basic | IPC-Net [53] | D | - | **93.7** | - | 86.7 | - |
| | SPHNet [54] | R | - | 87.1 | - | - | - |
| | RI-GCN [55] | R | - | 91 | - | - | - |
| | PointWOLF [42] | Patch-based (T+R+S+D+J) | PointNet++ [46] | 93.2 | - | **89.7** | 84.1 |
| | PatchAugment [56] | Patch-based (T+R+S+D+J) | DGCNN [43] | 93.1 | 95.6 | 86.9 | 79.7 |
| | | | PointNet++ [46] | 93 | 95.6 | 85.7 | 81 |
| | AdaPC [57] | Auto optimization (T+R+S+J) | - | 91.61 | - | 81.75 | - |
| | PointAugment [58] | Auto optimization (T+R+S+J) | DGCNN [43] | 93.4 | **96.7** | *83.1* | *76.8* |
| | Pointnet++ [46] | T+R | - | 91.9 | - | *84.3* | *77.9* |
| | PointNet [45] | R+J | - | 89.2 | - | *79.2* | *68.0* |
| | MaskPoint [47] | T+S | - | - | - | **89.7** | **84.6** |
| | CrossPoint [49] | T+R+S+F+D+J | DGCNN [43] | 91.2 | - | 81.7 | - |
| Specialized | Luo et al. [59] | Mixup | - | 87.6 | 94.2 | - | - |
| | PC-GAN [60] | Mixup | - | *84.5* | *95.4* | - | - |
| | Achlioptas et al. [61] | Mixup | - | 84.5 | 95.4 | - | - |
| | Point MixSwap [62] | Mixup | DGCNN [43] | 93.5 | 96 | 88.6 | - |
| | PointMixup [63] | Mixup | DGCNN [43] | *93.1* | *95.1* | *87.6* | *80.6* |
| | RSMix [64] | Mixup | DGCNN [43] | 93.5 | 95.9 | - | - |
| | SageMix[65] | Mixup | DGCNN [43] | 93.6 | - | 88 | 83.6 |
| | PointCutMix [66] | Mixup | PointNet++ [46] | 93.4 | 96.3 | 88.47 | 82.8 |
| | FMix [67] | Mixup | PointNet [45] | - | 89.57 | - | - |
| | MaxUp [68] | Adversarial deformation | - | 92.4 | - | - | - |

# Reference Used in Appendix A